\documentclass[a4paper,conference]{IEEEtran}

\usepackage{cite}

\usepackage[pdftex]{graphicx}
\graphicspath{{./images/}}
\DeclareGraphicsExtensions{.pdf,.jpeg,.png}

\usepackage{amsmath, amssymb}
\usepackage[utf8]{inputenc} 
\usepackage[T1]{fontenc}
\usepackage[ruled,vlined]{algorithm2e}
\usepackage{todonotes}

\newcommand\figwidth{\@empty}

\ifCLASSOPTIONcompsoc
 \usepackage[caption=false,font=normalsize,labelfont=sf,textfont=sf]{subfig}
\else
 \usepackage[caption=false,font=footnotesize]{subfig}
\fi
\usepackage{hyperref}
\usepackage{xcolor}
\hypersetup{
    colorlinks,
    linkcolor={red!50!black},
    citecolor={blue!50!black},
    urlcolor={blue!80!black}
}

\usepackage{xspace}

\newcommand{\ssep}{\; \middle | \;}
\newcommand{\R}{\mathbb{R}}
\newcommand{\PR}{\mathbb{P}}
\newcommand{\FF}{\mathcal{F}}
\newcommand{\JJ}{\mathcal{J}}
\newcommand{\OO}{\mathcal{O}}
\newcommand{\LL}{\mathcal{L}}
\newcommand{\Loss}{\mathbb{L}}
\newcommand{\pp}{\begin{pmatrix} p_1 & p_2 \end{pmatrix}}
\newcommand{\pphat}{\begin{pmatrix} \hat{p_1} & \hat{p_2} \end{pmatrix}}
\newcommand{\wall}{\textit{wall}}
\newcommand{\ceiling}{\textit{ceiling}}
\newcommand{\floor}{\textit{floor}}
\newcommand{\door}{\textit{door}}
\newcommand{\window}{\textit{window}}
\newcommand{\jproper}{\textit{proper}}
\newcommand{\jfalse}{\textit{false}}
\newcommand{\jinvalid}{\textit{invalid}}

\newcommand{\sAP}{\text{sAP}}
\newcommand{\jAP}{\text{jAP}}
\newcommand{\msAP}{\text{msAP}}
\newcommand{\mjAP}{\text{mjAP}}

\newcommand{\eg}{e.g.\@\xspace}
\newcommand{\etal}{\textit{et al.\@\xspace}}

\begin{document}
%
\title{Semantic Room Wireframe Detection\\ from a Single View}

\author{\IEEEauthorblockN{David Gillsjö, Gabrielle Flood and Kalle Åström}
\IEEEauthorblockA{Centre for Mathematical Sciences,
 Lund University, Sweden\\
\{david.gillsjo,gabrielle.flood,karl.astrom\}@math.lth.se
}}
%

\maketitle

\begin{abstract}
  Reconstruction of indoor surfaces with limited texture information or with repeated textures,
a situation common in walls and ceilings, may be difficult with a monocular Structure from Motion system.
We propose a Semantic Room Wireframe Detection task to predict a Semantic Wireframe from a single perspective image.
Such predictions may be used with shape priors to estimate the Room Layout and aid reconstruction.
To train and test the proposed algorithm we create a new set of annotations from the simulated Structured3D dataset.
We show qualitatively that the SRW-Net handles complex room geometries better than previous Room Layout Estimation algorithms
while quantitatively out-performing the baseline in non-semantic Wireframe Detection.

\begin{figure}
  \centering
    \includegraphics[width=0.45\linewidth]{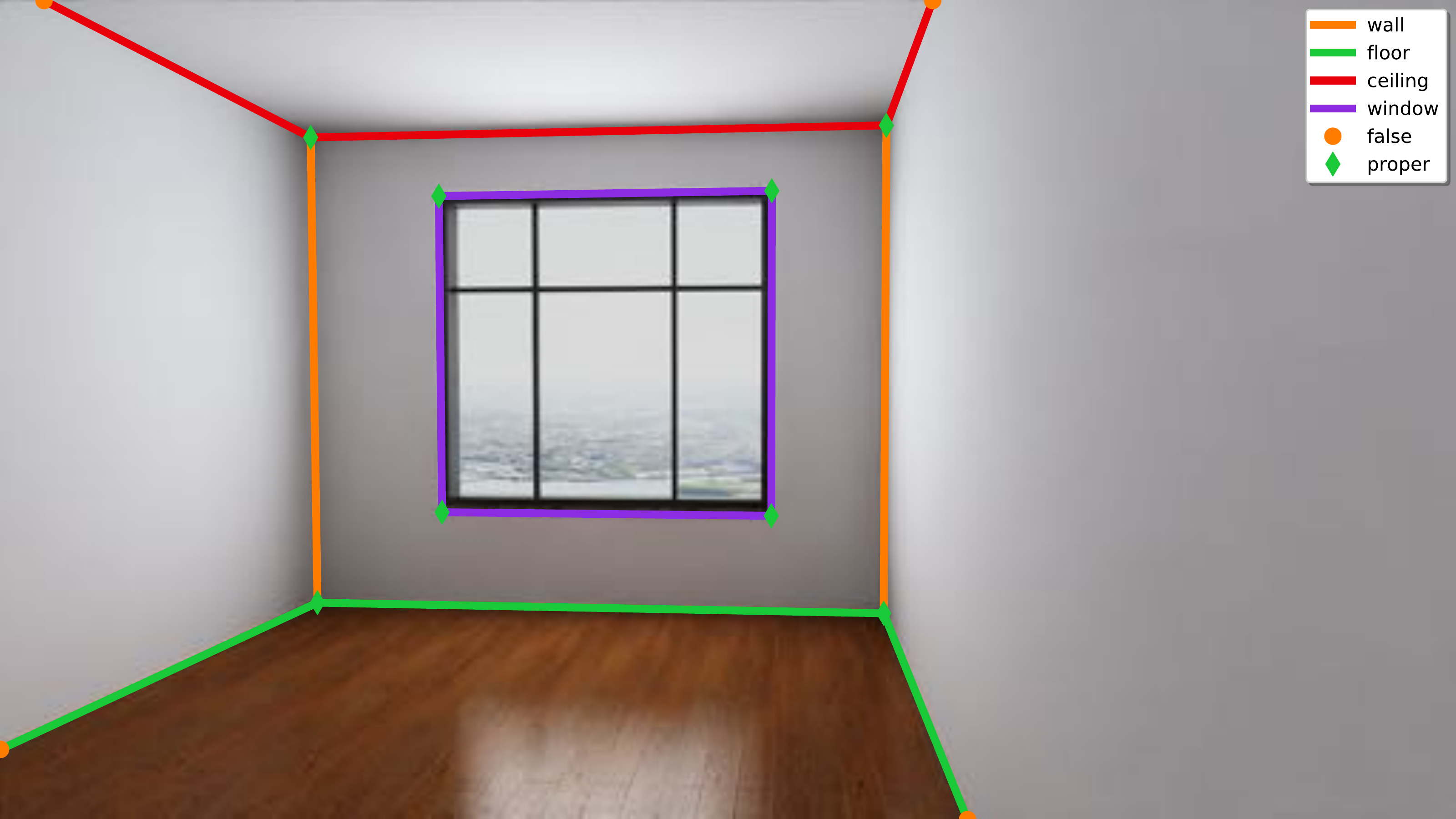}
    \includegraphics[width=0.45\linewidth]{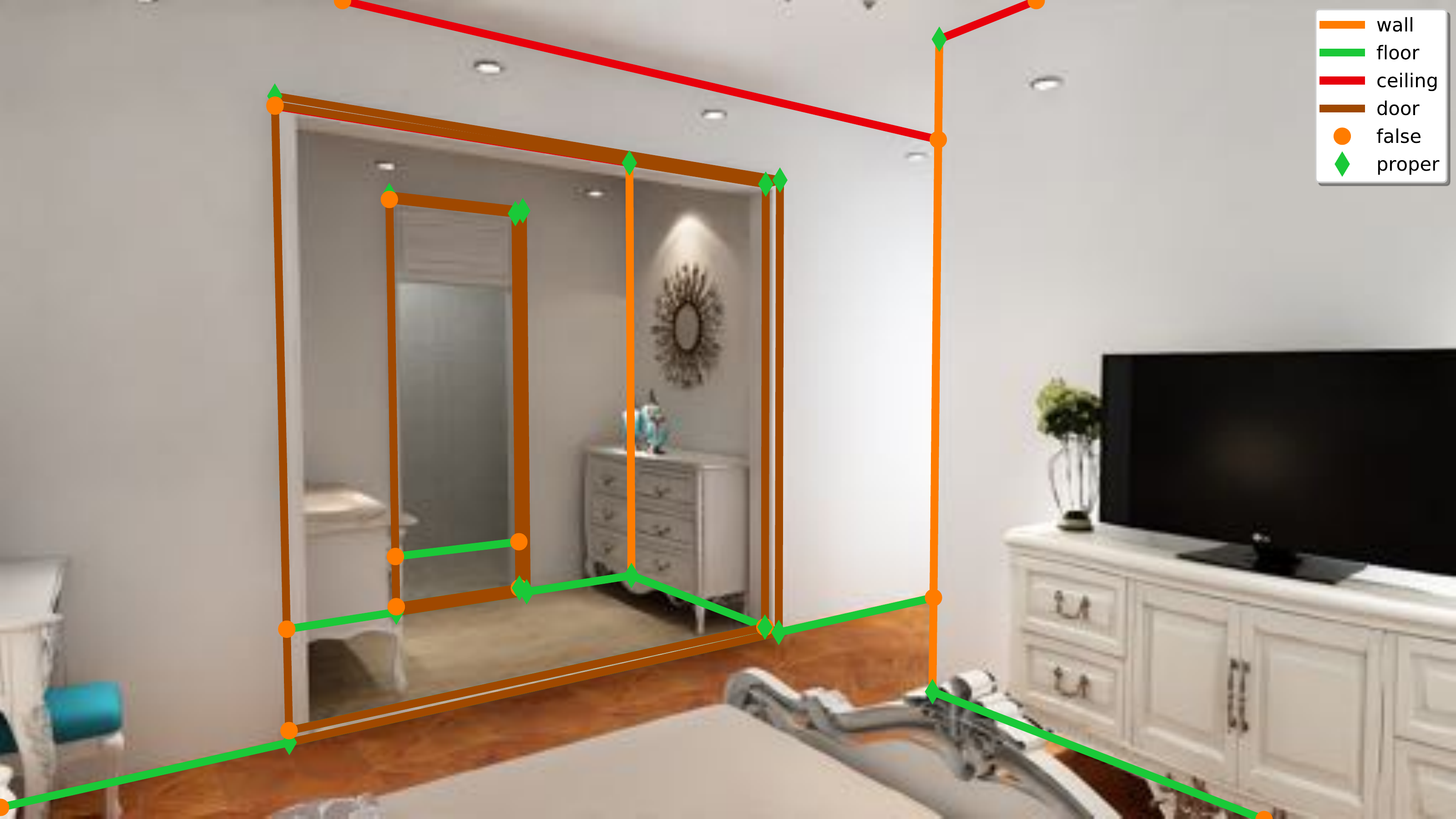}\\
    \includegraphics[width=0.45\linewidth]{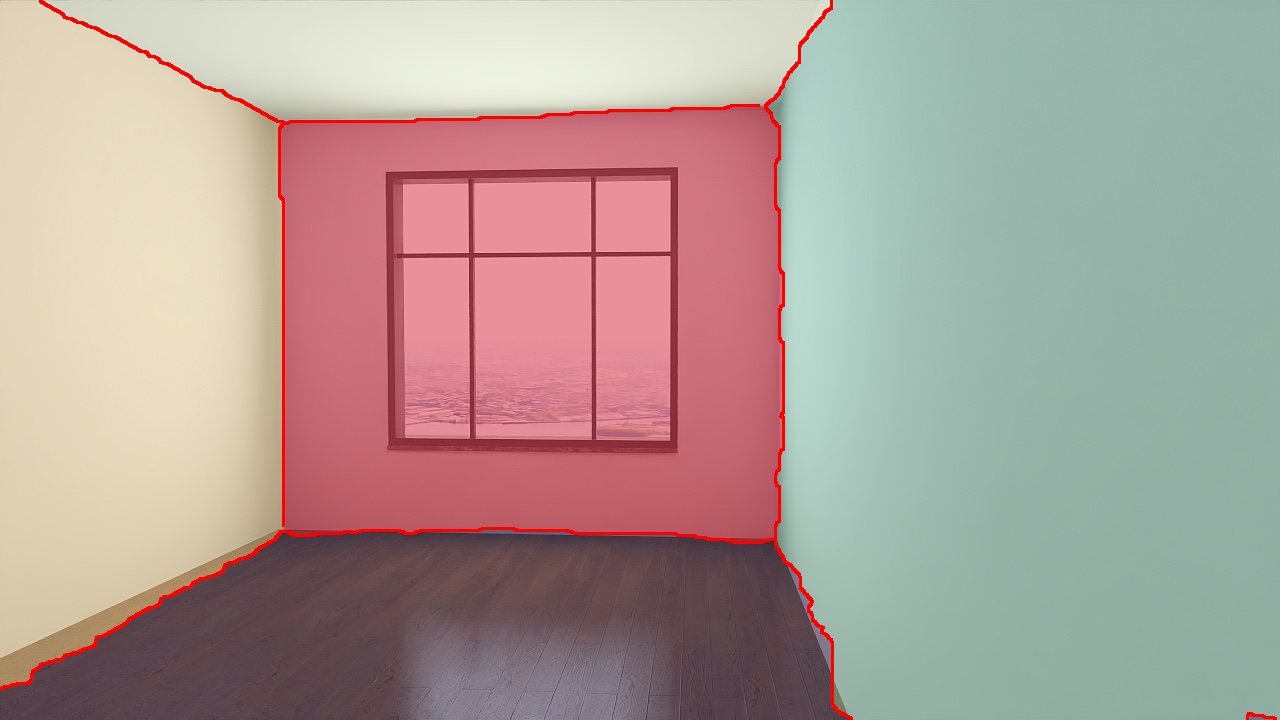}
    \includegraphics[width=0.45\linewidth]{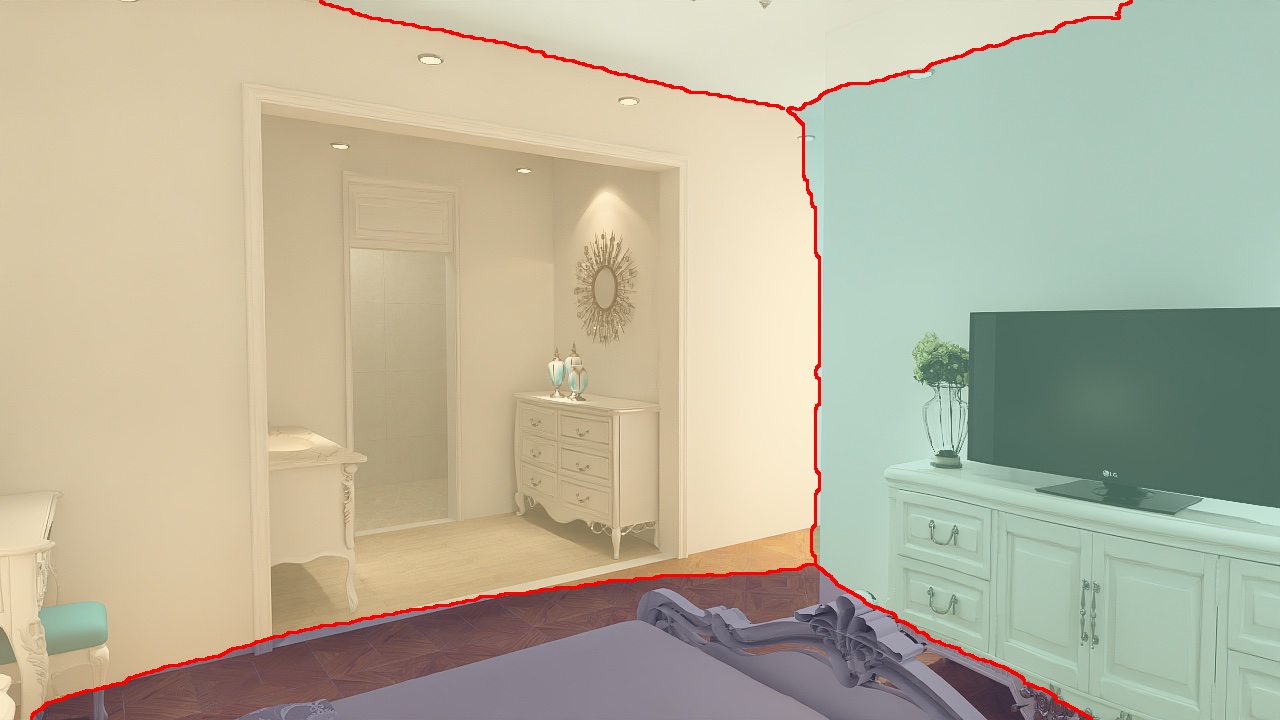}
  \caption{The images on top show the desired results from Semantic Room Wireframe detection and below are the results from a Room Layout estimation algorithm \cite{lin2018layoutestimation}. Lines are added between all detected planes for comparison. The algorithm produces satisfying results on simple scenes (left) but is lacking on more complex scenes, \eg to the right where the wide opening is erroneously classified as part of the wall.}
  \label{fig:motivation}
\end{figure}

\end{abstract}


%
\IEEEpeerreviewmaketitle

\section{Introduction}
\label{sec:intro}
Reconstruction of indoor spaces is useful for many applications, \eg Virtual Reality, Real Estate sales and navigation for robots.
Using consumer grade devices with a single camera (and possibly IMU) is preferred since this enables the application on a larger number of platforms.
For this type of reconstruction one typically uses Structure from Motion (SfM) or Monocular Vision SLAM (Simultaneous Localization and Mapping).
These rely on point features -- points in the image with high saliency in terms of texture.
This may be problematic in indoor spaces since most walls and ceilings are without salient textures.

This has been the inspiration for research on Room Layout Estimation,
which is the task of estimating the Room Geometry in terms of planar surfaces.
Often strong priors about the relation between the surfaces are used and Room Geometries are assumed to be too simple, see Figure \ref{fig:motivation}.
Furthermore the multi-view case is often handled using Panorama, which is not always applicable.

To relax the assumptions and handle multi-view scenarios we believe that the way forward is to develop several pipelines:
semantic segmentation in the image, semantic segmentation in 3D, image object detection, corner detection and semantic line detection.
These could then be used together -- ideally in an end-to-end fashion. In this work our aim is to concentrate on one of these aspects and propose a
Semantic Room Wireframe Detection task, where connected semantic line segments forming a wireframe, are to be detected.
Since this is a detection task no strong assumptions are made and the detections can be used in a larger optimization framework to handle more complex Room Geometries captured from multiple views.
Compared to previous methods we for example include labels for neighbouring rooms visible through open doors and label doors and windows in addition to the room structure.

We generate annotations, which consist of semantic line segments and semantic junctions from the synthetic Structured3D dataset.
To perform the detection we adapt HAWP \cite{HAWP}, which is a CNN (Convolutional Neural Network) developed for the Wireframe Detection task, and extend it with a Graph Convolutional Network (GCN) \cite{KipfW17GCN} module.
We call the new network SRW-Net (Semantic Room Wireframe Network).
We show qualitatively that the SRW-Net\footnote{See \url{https://github.com/DavidGillsjo/SRW-Net} for code and data.} handles complex room geometries better than previous Room Layout Estimation algorithms while quantitatively out-performing HAWP \cite{HAWP} and L-CNN \cite{zhou2019end} for non-semantic Wireframe Detection.
\noindent
\textbf{The contributions of this paper are:}
\begin{itemize}
  \item Proposal of the Semantic Room Wireframe detection task.
  \item New annotations for this task which we call Structured3D-SRW.
  \item Implementation and evaluation of a Neural Net for Semantic Room Wireframe detection, SRW-Net.
\end{itemize}

\section{Related Work}
Room Layout Estimation was studied early by Hedau \etal \cite{Hedau} which used the Manhattan World assumption \cite{manhattan} as prior for the room shape.
Early methods \cite{prenetwork1,prenetwork2,prenetwork3} used a pipeline which first extracted handcrafted features, then did vanishing point detection and lastly hypothesis generation and ranking. Recent approaches use CNNs, \eg Mallya and Lazebnik \cite{Mallya} use structured edge detection forests, and a CNN to predict edge probability masks. Lin \etal \cite{lin2018layoutestimation} propose an end-to-end CNN with adaptive edge penalty and smoothness terms for pixelwise segmentation of the room image.
RoomNet \cite{lee2017roomnet} directly predicts ordered keypoints in a room layout and  DeepRoom3D \cite{deeproom} use an end-to-end CNN to predict a cuboid.
Most recent methods \cite{cfile,zhao2017physics,Yan,doubleref2020} use CNNs to predict edges and then optimize for the Room Layout using geometric priors.
The datasets mostly used are LSUN \cite{yu2016lsun} and Hedau \cite{Hedau}.

Zhang \etal \cite{panocontext} advocated the use of Panorama images for increased performance in Room Layout Estimation. This has since been an inspiration for Panoramic Room Layout Estimation, which allows for better use of context and prior knowledge about camera orientation and calibration.
Algorithms are typically based on CNNs \cite{zou2018layoutnet, yang2019dula,sun2019horizonnet} using different image representations and post-processing.
For example CFL \cite{fernandez2020corners} predicts the 3D layout from a spherical image using edge and corner maps.
Despite the promising results from these methods Room Layout Estimation from panorama does, indeed, limit the use of more conventional images.
Common datasets  for panorama are SUN360 \cite{SUN360},  Stanford (2D-3D-S) \cite{standford2D3D} and Structured3D \cite{Structured3D}.

Recently, general Room Layout Estimation from a single perspective image, which does not assume cuboid shape, has been studied.
\cite{general3d2020} solves a discrete optimization problem over 3D polygons using both RGB and Depth information.
\cite{HowardJenkins2018ThinkingOT} use plane detection to form a 3D model over a video sequence.
\cite{NonCuboidRoom2022} use a combination of plane, depth and vertical line detections to estimate a general Room Layout.
Neither of these handle detections through open doorways.
Data for this task is Structured3D \cite{Structured3D} and ScanNet \cite{dai2017scannet}.

There is a vast literature on the use of lines for 3D understanding, even though SfM and SLAM are most often used in terms of points. For example understanding consistency constraints \cite{heyden1994consistency}, exploiting such constraints for 3D understanding and SfM methods using only lines \cite{kahl1998structure,astrom1999structure,oskarsson2004minimal,bartoli2005structure,larsson2017efficient,micusik2017structure,lemaire2007monocular}.

Detection of connected line segments has been studied as the task of Wireframe Estimation \cite{xia2014accurate,xue2019learningcvpr,xue2019learningpami, HAWP}.
However, these methods do not try to use semantic understanding of the lines and junctions and do not exploit such semantic understanding in terms of Room Layout Estimation. While recent works mostly form the Wireframe using junction proposals some works use line predictions directly.
For example in \cite{lgnn2020}, a combination of line predictions and graph networks are used.
YorkUrban \cite{YorkUrban} and ShanghaiTech \cite{ShanghaiTech} are common datasets. Other approaches that predict semantic information for line segments \cite{sun2019semlsd} are focused around object detection rather than detection of indoor room structures.

\section{Semantic Room Wireframe Detection}
We define a Semantic Room Wireframe Detection task. The prediction is done from a single perspective image and should result in a number of semantic lines in the image which -- when connected by junctions -- form a wireframe marking the intersections between the planes defining the room layout.

The task can be seen as a combination of Room Layout Estimation and Wireframe Detection. It is more flexible than Room Layout Estimation since no priors on the room structure are used but also contains more information than regular Wireframe Detection.  The results can, \eg, be used to aid reconstructions, s.a.\ SfM and SLAM where salient textures are missing.
It is also suitable for multi-view Room Layout Estimation, where it gives more flexibility for global optimization and better information in door openings.
For an example of the desired output of a Semantic Room Wireframe Detector, see the top row of Figure \ref{fig:motivation}.

\section{Structured3D-SRW}
\label{sec:dataset}
Our annotations for this task, which we call Structured3D-SRW, are generated from Structured3D \cite{Structured3D} which is a large-scale, photo-realistic simulated dataset with 3D structure annotations. It consists of 3500 scenes, with a total of 21'835 rooms and 196'515 frames.
Structured3D contains 3D information about the room structure and is built up from junctions, lines and plane polygons.
A junction is a point in $\R^3$ and a line is defined as the line segment formed by two junctions.
A plane polygon is defined by a set of connected line segments and its parameters $\pi$ defining the normal vector and plane equation.
These planes are labeled with semantic information, for example if the plane is a floor and part of a kitchen.

There are both panoramic and perspective images in the dataset. For each room, there are configurations both without furniture, with a little and with a lot of furniture. For each of the images, there is also a depth map and a semantic mask. The data is split so there are 3000 training, 250 validation and 250 testing scenes.

\subsection{Data Analysis}
\label{sec:analysis}
Our annotations are generated by combining the 3D information, semantic masks and perspective RGB images with full furnishing.
For the images with full furnishing there are 202 scenes missing from the training set, so that leaves us with 2'798 training scenes.
While working with the data, we identified three main issues.

\begin{figure}
  \centering
  \includegraphics[trim=0 165 0 30, clip, width=\linewidth]{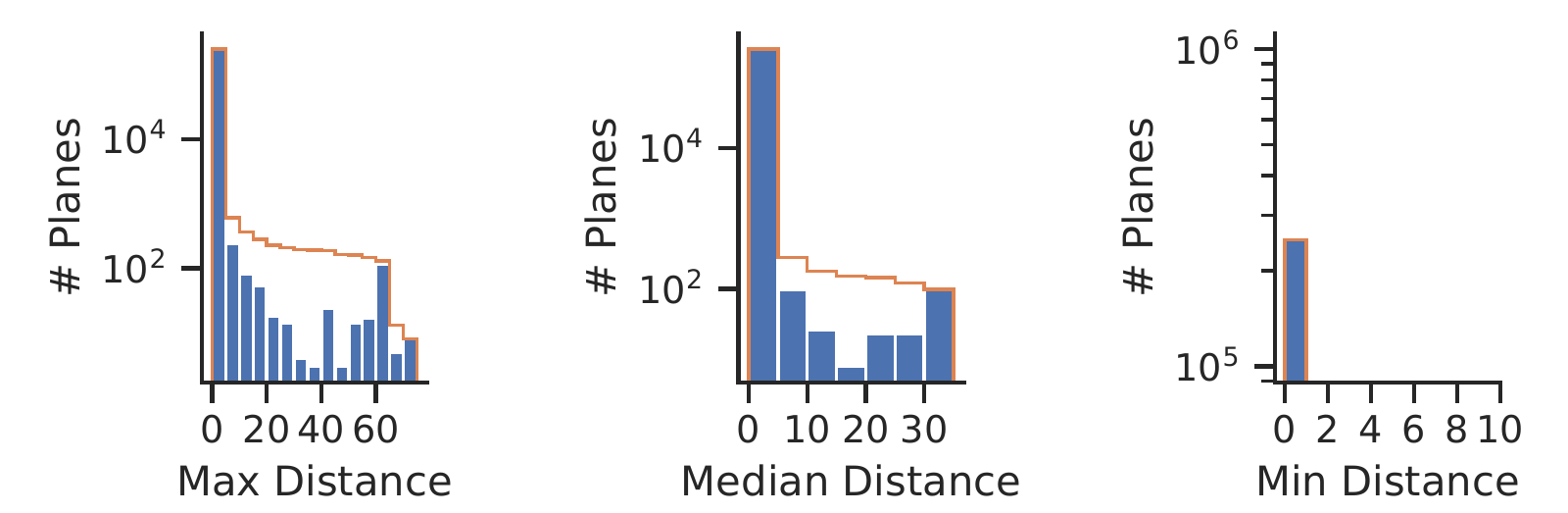}%
  \caption{Histograms for the maximum, median and minimum distance from each plane to its junctions using the supplied plane equation.
  Note that the plane parameters are not optimized to minimize distance to all junctions.}
  \label{fig:wall_hist}
\end{figure}

Firstly, there are a few planes to which only two junctions belong. However, only two junctions cannot uniquely define a plane.
Secondly, it turns out that there are a number of planes for which the defining junctions do not lie in the plane.
We calculated the Euclidean distance between all plane junctions and the plane according to supplied parameters and the result can be seen in Figure \ref{fig:wall_hist}.
We see that there are around $10^3$ planes with a maximum distance larger than 5mm. This is problematic since the assumption is that all surfaces are planar in this synthetic dataset.
Thirdly, there is no information about whether the doors in Structured3D are open or closed. Since one of the strengths of our proposed method is that more complex structures can be captured, we want the method to identify open doors to create a wireframe also in the room behind them. One example of this can be seen in Figure \ref{fig:junction_example}, where the large opening in the left wall is annotated as a door. Knowing that it is open, we can train our method to detect the structure of the second room as well, as annotated in the image.

We have implemented the following measures to adjust for these three main issues:

\textbf{(i)}
For the walls with only two junctions, we cannot form plane polygons to make occlusion checks. Because of this we simply discard the scenes containing such planes.

\textbf{(ii)}
To verify that junctions lie in the plane, we chose to estimate new plane parameters from the junctions by solving the DLT problem
\begin{align}
  \min_{\| \pi \|^2=1} \| M\pi \|^2, \; \text{for} \quad
  M &=\begin{pmatrix} p_1 & p_2 & \hdots & p_n \end{pmatrix}^T, \\
  p_i &= \begin{pmatrix} x_i & y_i & z_i & 1 \end{pmatrix}^T,
\end{align}
where $p_i$ is a junction coordinate and $\pi$ are the four plane parameters s.t.\ a point $q$ is in the plane iff. $q^T \pi = 0$.
With these plane parameters we get lower maximum and median distances, compared to using the supplied parameters.
Still, there are many planes that have large distances to their junctions. We therefore remove any scene with planes that have maximum error larger than 1mm.
\textbf{(i)} and \textbf{(ii)} combined leaves us with 2'502, 218 and 205 scenes corresponding to 57'252, 5'684 and 5'085 images for training, validation and testing, respectively.

\textbf{(iii)}
We do the following to determine whether doors are open or closed:
For each door polygon $D_{i,j}$ belonging to wall plane $\pi_i$ we take a set of 100 uniformly distributed sample points $\hat{D}_{i,j} = \{ p_k \in D_{i,j} \}$.
For each sample set $\hat{D}_{i,j}$ and each perspective image $\hat{I}_l$ for a scene, we find the sample points
\begin{equation}
  \hat{D}^l_{i,j} = \left \{ (x_k,y_k, 1)^T \sim P_l p_k \ssep ( x_k,y_k ) \in \hat{I}_l \right \},
\end{equation}
in pixel coordinates using the camera matrix $P_l = K_l[R_l \; t_l]$, where $K_l$ are intrinsic parameters, $R_l$ rotation matrix and $t_l$ translation vector.
We then compute a door closed ratio
\begin{equation}
c_{i,j} = \frac{
	\sum_l \sum_k \alpha_{lk}
}{
	\sum_l \sum_k 1
} , \qquad
\alpha_{lk} = \begin{cases}
1, \, \text{if} \; \hat{I}_l[y_k,x_k] =  \door{}, \\
0,  \, \text{otherwise},
\end{cases}
\end{equation}
and say that the door $\hat{D}_{i,j}$ is closed if $c_{i,j} > 0.3$.

\subsection{Semantic Information}
\label{sec:semantic_info}
In Structured3D each plane is annotated with a type and a semantic class.
By combining the information in these two fields we mark each plane as either \wall{}, \floor{}, \ceiling{}, \door{} or \window{}.

We choose to have one label per line and define each line label as the combination of the two planes it belongs to.
If we look at the last column of Table \ref{tab:line_labels} we see that there are seven occurring classes, where \door{} and \window{} planes yield two different line labels each.
Since most of the \door{}-\door{} and \window{}-\window{} lines will be concealed behind doors and window glass we choose to merge these
with \door{}-\wall{} and \window{}-\wall{}, respectively. We then get the mapping in Table \ref{tab:line_labels},
which consists of the same five labels as for the planes, namely \wall{}, \floor{}, \ceiling{}, \door{} and \window{}.

\begin{figure}
     \centering
     \includegraphics[width=\linewidth]{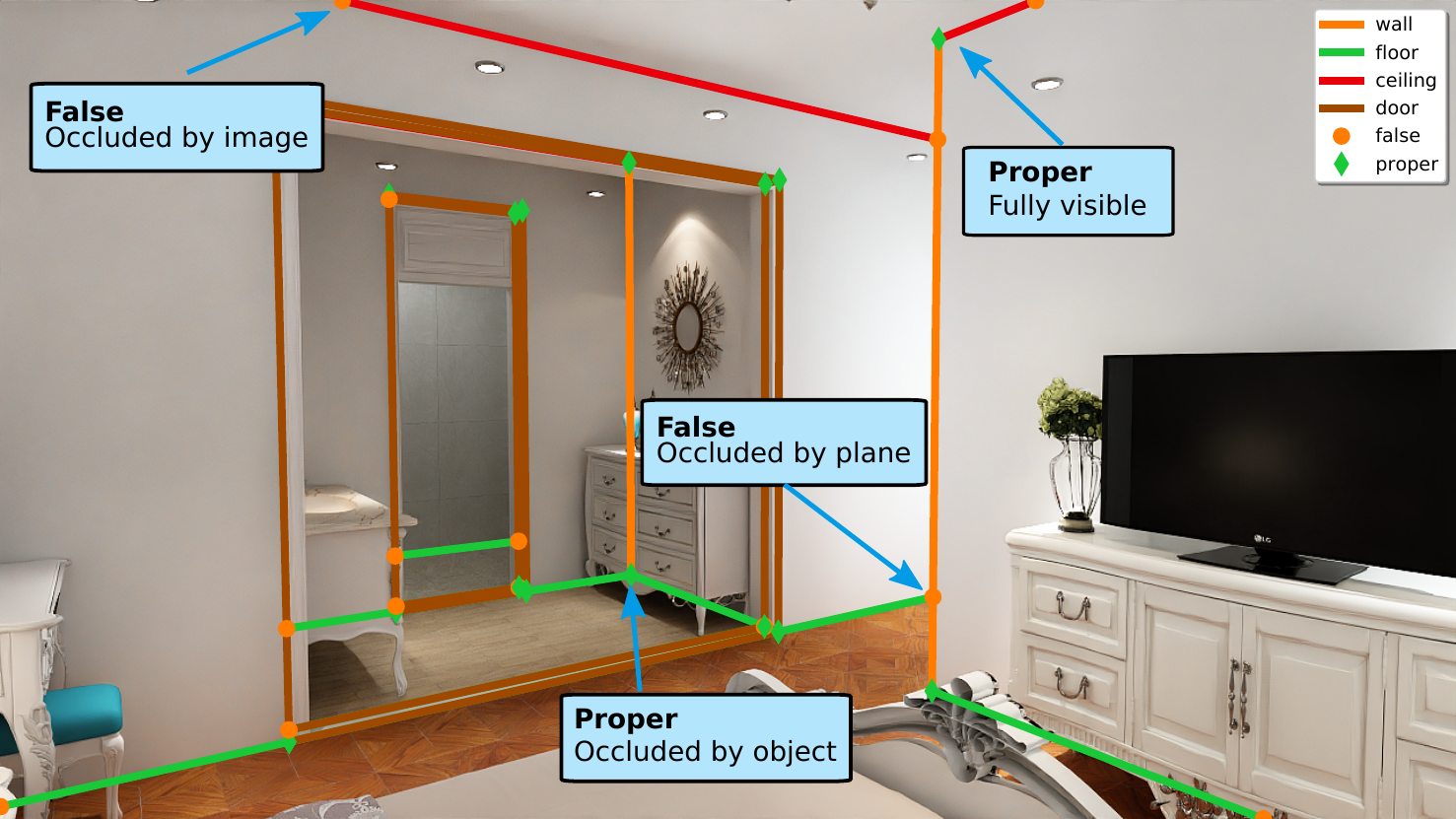}
     \caption{Example cases for junction labels \jproper{}  and \jfalse{} .}
     \label{fig:junction_example}
\end{figure}

Furthermore, we also label the junctions. The dataset consists of two different kinds of junctions.
There are junctions corresponding to a 3D junction where three planes meet; these we call \jproper{} and they should be detected even when occluded by furniture.
There are also junctions which occur either due to the camera having limited field of view, or due to occlusion from other planes.
These we call \jfalse{}.
See Figure \ref{fig:junction_example} for an example.

\begin{table}
  \renewcommand{\arraystretch}{1.2}
  \caption{Mapping of line labels from plane labels.}
  \label{tab:line_labels}
  \begin{center}
  \begin{tabular}{|l|c|c|}
  \hline
  Line label & Plane labels & Occurrence [\%]\\
  \hline\hline
  \door{} & \door{}-\wall{} & 20.4 \\
  \wall{} & \wall{}-\wall{} & 15.5 \\
  \ceiling{} & \ceiling{}-\wall{} & 15.5 \\
  \floor{} & \floor{}-\wall{}  & 15.5 \\
  \door{} & \door{}-\door{} & 11.6 \\
  \window{} & \window{}-\wall{} &  11.0 \\
  \window{} & \window{}-\window{} & 10.4 \\
  \hline
  \end{tabular}
  \end{center}
\end{table}

\section{SRW-Net}
\begin{figure*}[!t]
     \centering
     \includegraphics[width=0.8\linewidth]{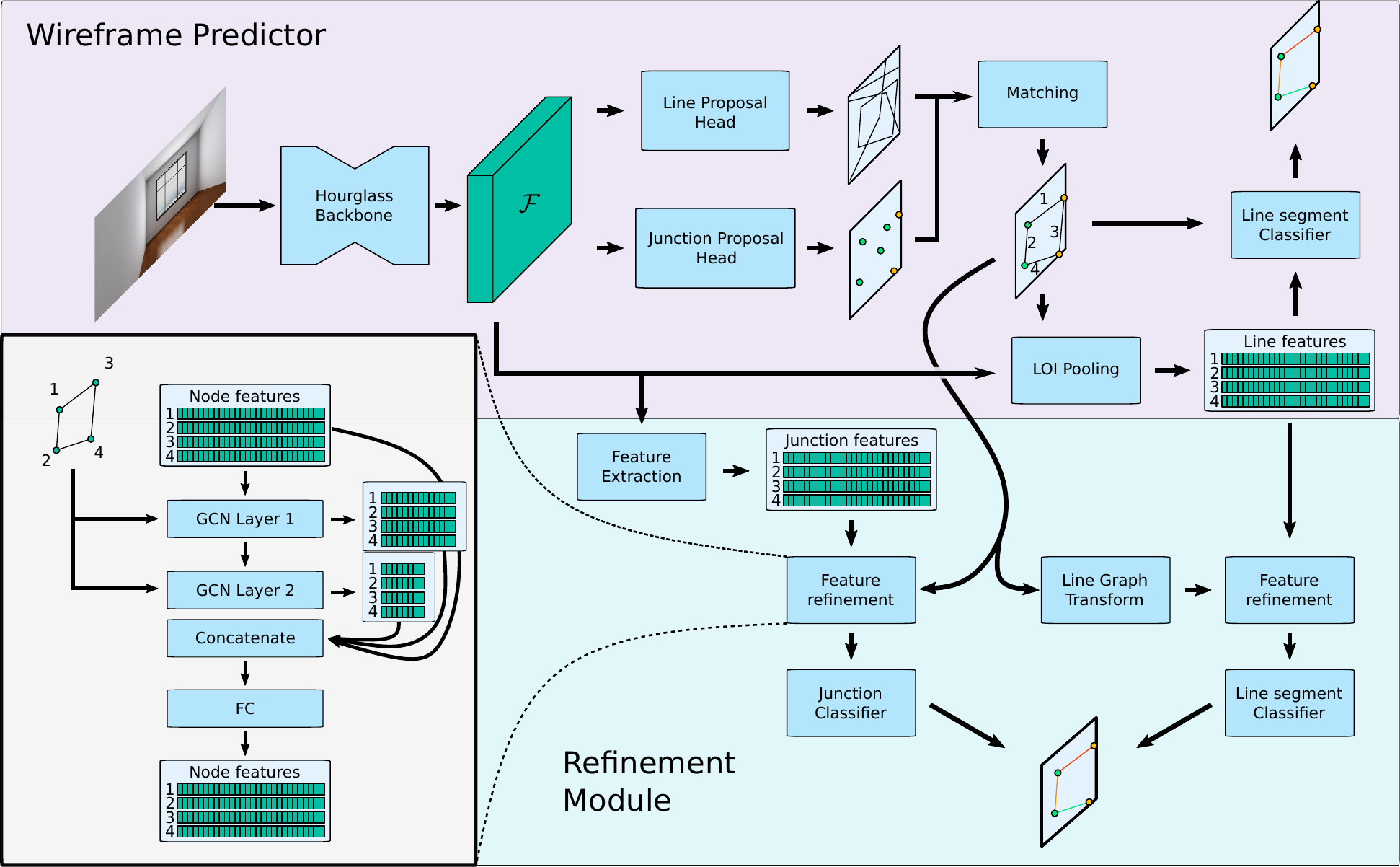}
     \caption{An overview of the SRW-Net architecture.}
     \label{fig:network_arch}
\end{figure*}

For the CNN architecture we chose to base it on HAWP (Holistically-Attracted Wireframe Parsing), presented in Xue \etal \cite{HAWP}.
Since our data contains semantic labels for both lines and junctions, but fewer lines per image, we had to make some adjustments to the architecture and parameters.
In addition to the changes we introduce a GCN module, trained separately, to refine the features using neighbours in the wireframe.
See Figure \ref{fig:network_arch} for an overview of the architecture.

\subsection{Backbone}
Before fed into the network, the image is resized to $H \times W$ pixels, where $H=W=512$. It is then passed through a backbone Hourglass Network \cite{hourglass} which generates a latent feature space $\FF$ of size $N \times \hat{H} \times \hat{W}$, where $\hat{H}=\hat{W}=128$ and $N=128$ is the number of channels.
These features are then used by the different network heads.

\subsection{Junction Proposals}
The junction proposal head predicts from $\FF$ a junction label matrix $\JJ$ of size $3 \times \hat{H} \times \hat{W}$
with one-hot encoding for labels \jinvalid{}, \jfalse{}, \jproper{}. It also predicts an offset matrix $\OO$ of size $2 \times \hat{H} \times \hat{W}$.
They are computed using a $1 \times 1$ convolutional layer and then sigmoid for $\OO$ and softmax for $\JJ$.
This is different from HAWP since they did not have labels for junctions and used sigmoid for $\JJ$.
We then form $\JJ_v = 1-\JJ [ \jinvalid{} ] $.
After a $3 \times 3$ NMS (Non-Maximum Suppression) to $\JJ_v$ the top $K$ bins $\JJ_K =  \{ (x_k,y_k), k=1...K \}$ are selected ($K=80$)
and the offset map is used to compute the final position of the junctions $\JJ_f = \{ (x_k,y_k) + \mathcal{O}(x_k,y_k) \cdot w, k=1...K \}$
where $w$ is a rescaling factor. Like HAWP we use cross entropy loss to form $\Loss(\JJ,\hat{\JJ})$  and L1 loss for $\Loss(\OO,\hat{\OO})$.
But we weight $\jinvalid{}$ as 1:250 in $\Loss(\JJ,\hat{\JJ})$ to account for bias and improve recall.
The final loss function is
$\Loss_j = \lambda_{msk} \cdot \Loss(\JJ,\hat{\JJ}) + \lambda_{off} \cdot \JJ_v \odot \Loss(\OO,\hat{\OO})$,
where $\odot$ is elementwise multiplication, $\lambda_{msk}$ and $\lambda_{off}$ are design parameters.

\subsection{Line Segment Proposal}
The line segment proposal head is the same as for HAWP and is a modification of AFM \cite{xue2019learningcvpr} which uses a parametrization of line segments using four parameters.
It is, however, not unique and each parametrization may yield up to three line segments.
The proposal head takes $\FF$ and predicts a $4 \times \hat{H} \times \hat{W}$ matrix, where the first dimension is the 4D AFM parametrization, which is then converted to
the endpoint line segment representation. The loss function is the same as for HAWP and we denote it $\Loss_{LS}$.

\subsection{Line Segment and Junction Proposal Matching}
Now we have line segment proposals and junction proposals.
Each line segment's endpoints are matched to their closest junction.
If the Euclidean distance between each endpoint and matched junction is below $\tau=10$, we form a line segment between these two junctions.
We denote the set of matched line segments $\LL_m$.

\subsection{Matched line segment classification}
Each matched line segment $l \in \LL_m$ will be classified as either
\jinvalid{}, \wall{}, \floor{}, \ceiling{}, \window{} or \door{}.
Following the implementation of HAWP we use the LoIPool operation \cite{zhou2019end} from L-CNN.
First $s=32$ uniformly spaced points along the line are sampled from $\FF$.
All features are then reduced, concatenated and given to the 2-layer fully connected classifier ending with the softmax operation and cross entropy loss $\Loss_{cls}$.
This is trained end to end and the final loss is $\Loss = \Loss_{junc} + \Loss_{LS} + \Loss_{cls}$.
During training we balance negative and positive examples while also providing hard negative examples from the ground truth data, see HAWP \cite{HAWP} for details.
We reduced the number of sampled lines to 100 negative and 100 positive.

\subsection{Wireframe Refinement - Graph Convolutional Network}
This GCN module uses the graph structure to augment the line and junction features with encoded features from the neighbouring lines and junctions prior to classification.
The module is trained separately from the larger network and takes lines with \jinvalid{} score less than $0.95$ as input.
We form two graphs, the junction graph is simply the estimated wireframe where junction features are nodes and lines define edges.
The line segment graph is created by taking the Line Graph \cite{BEINEKE_linegraph} of the junction graph, s.t. the line segment features now are nodes.
The architecture of the GCN layers are encoder-based with skip-connections to a fully connected classifier layer, see Figure \ref{fig:network_arch}.
We use 4 GCN layers for junction and line head respectively. To achieve good results we sample positive and negative examples as 5:1 for the loss function. The loss and classifier architecture are the same as previously.

\section{Evaluation}
\textbf{sAP} (structural average precision) \cite{zhou2019end} is the area under the precision and recall curve for a set of scored detected line segments.
Let $\hat{L} = \pphat \in \LL_g$ be any line segment in the set of true line segments.
A detected line segment $L = \pp$ is a true positive if
\begin{equation}
  \label{eq:sAP}
  \min_{\hat{L} \in \LL_g} \delta(L,\hat{L}) = \| p_1 - \hat{p}_1 \|^2_2 + \| p_2 - \hat{p}_2 \|^2_2 \leq \beta,
\end{equation}
where $\beta$ is a design parameter. We also require that only one detection is matched to each true line segment. Any extra predictions are marked as false positives.
In our experiments we evaluate the metric at $\beta = 5,10,15$ at $128 \times 128$ resolution and denote the metrics
$\sAP^5$, $\sAP^{10}$ and $\sAP^{15}$ respectively.
We calculate sAP for each separate label and then take the mean to form $\msAP$.
We also calculate $\sAP^m = \sum_\beta \sAP^\beta /3$ for each label and take its mean across labels as $\msAP^m$.

\textbf{NMS} (Non Maximum Suppression) is also performed on line segments to improve performance.
We use the same distance $\delta$ as in Equation (\ref{eq:sAP}) and say that if a line segment $L$ has a neighbour $\hat{L}$ s.t.\
$ \delta(L,\hat{L}) < \gamma^2$ with the same predicted label and higher score, $L$ is removed. We use $\gamma = 3$.

\textbf{jAP} (junction AP), is analogous to $\sAP$.
Instead of the criteria in Equation (\ref{eq:sAP}) we take the Euclidean distance between the junction and the closest ground truth junction
of the same label. The thresholds are $0.5$, $1.0$ and $2.0$ and $\mjAP^m$, $\jAP^m$ follows as before.

\section{Experiments}
All experiments were done using pre-trained weights from HAWP as initialization for applicable layers, as we noticed this produced better results than training from scratch.
Models were trained for 40 epochs with learning rate $4\cdot10^{-4}$, weight decay $10^{-4}$ and batch size $11$ on a Nvidia Titan V.
The learning rate was reduced to  $4\cdot10^{-5}$ at epoch 25.
For the GCN refinement module we trained for 10 epochs with learning rate $1.2\cdot10^{-3}$, weight decay $10^{-4}$ and batch size $60$.
The learning rate was reduced to  $1.2\cdot10^{-4}$ at epoch 5.

\subsection{Model evaluation}
\label{sec:eval}
We evaluate the proposed model on the test set.
See Table \ref{tab:model_eval} for AP numbers and Figure \ref{fig:eval_PR} for the PR (precision recall) curve for threshold 10.
We see that \floor{} is by far the most difficult line type to detect correctly.
Since there is no method available for direct comparison we trained HAWP \cite{HAWP} and L-CNN \cite{zhou2019end} on Structured3D-SRW using their code and training methods.
All models are initiated with the pre-trained weights provided by the code releases.
Since these algorithms output a non-semantic wireframe we generate a wireframe prediction from our method by calculating the score as $(1-\jinvalid{})$ for junctions and lines.
In Table \ref{tab:model_nonsemantic_eval} we see that our method out-performs both HAWP and L-CNN in line segment detection (\sAP).
HAWP performs better in the junction metric \jAP.

\begin{table}
  \renewcommand{\arraystretch}{1.2}
  \caption{AP Scores for the final model.}
  \label{tab:model_eval}
  \centering
  \begin{tabular}{|l|r|r|r|r|}
  \hline
   Type      &   $\text{sAP}^{5}$ &   $\text{sAP}^{10}$ &   $\text{sAP}^{15}$ &   $\text{sAP}^{m}$ \\
  \hline
  \ceiling{} &               35.5 &                42.4 &                46.5 &               41.5 \\
  \door{}    &               44.7 &                49.6 &                52   &               48.8 \\
  \floor{}   &               15.5 &                21.6 &                24.6 &               20.6 \\
  \wall{}    &               39.8 &                47.3 &                51.2 &               46.1 \\
  \window{}  &               49.5 &                55.4 &                58.2 &               54.4 \\
  \hline
  $\text{mAP}$ &             37   &                43.3 &                46.5 &               42.3 \\
  \hline\hline
   Type         &   $\text{jAP}^{0.5}$ &   $\text{jAP}^{1}$ &   $\text{jAP}^{2}$ &   $\text{jAP}^{m}$ \\
  \hline
  \jfalse{}      &                 17.8 &               36   &               43.7 &               32.5 \\
  \jproper{}     &                 21.9 &               39   &               46.8 &               35.9 \\
   \hline
   $\text{mAP}$ &                 19.9 &               37.5 &               45.3 &               34.2 \\
  \hline
  \end{tabular}
\end{table}

\begin{table}
  \renewcommand{\arraystretch}{1.2}
  \caption{Comparision of the (1-\jinvalid{}) score against HAWP and L-CNN.}
  \label{tab:model_nonsemantic_eval}
  \centering
  \begin{tabular}{|l|r|r|r|r|}
  \hline
   Arch.    &  $\text{sAP}^{m}$          &   $\text{jAP}^{m}$ \\
  \hline
   LCNN    &                       31.8  &             34.9 \\
   HAWP    &                       46.0  &     \textbf{37.0} \\
   SRW     &               \textbf{46.6} &             36.2 \\
  \hline
  \end{tabular}
\end{table}

\renewcommand{\figwidth}{.48\linewidth}
\begin{figure*}[!t]
     \centering
     \includegraphics[trim=0 0 0 25, clip, width=\figwidth]{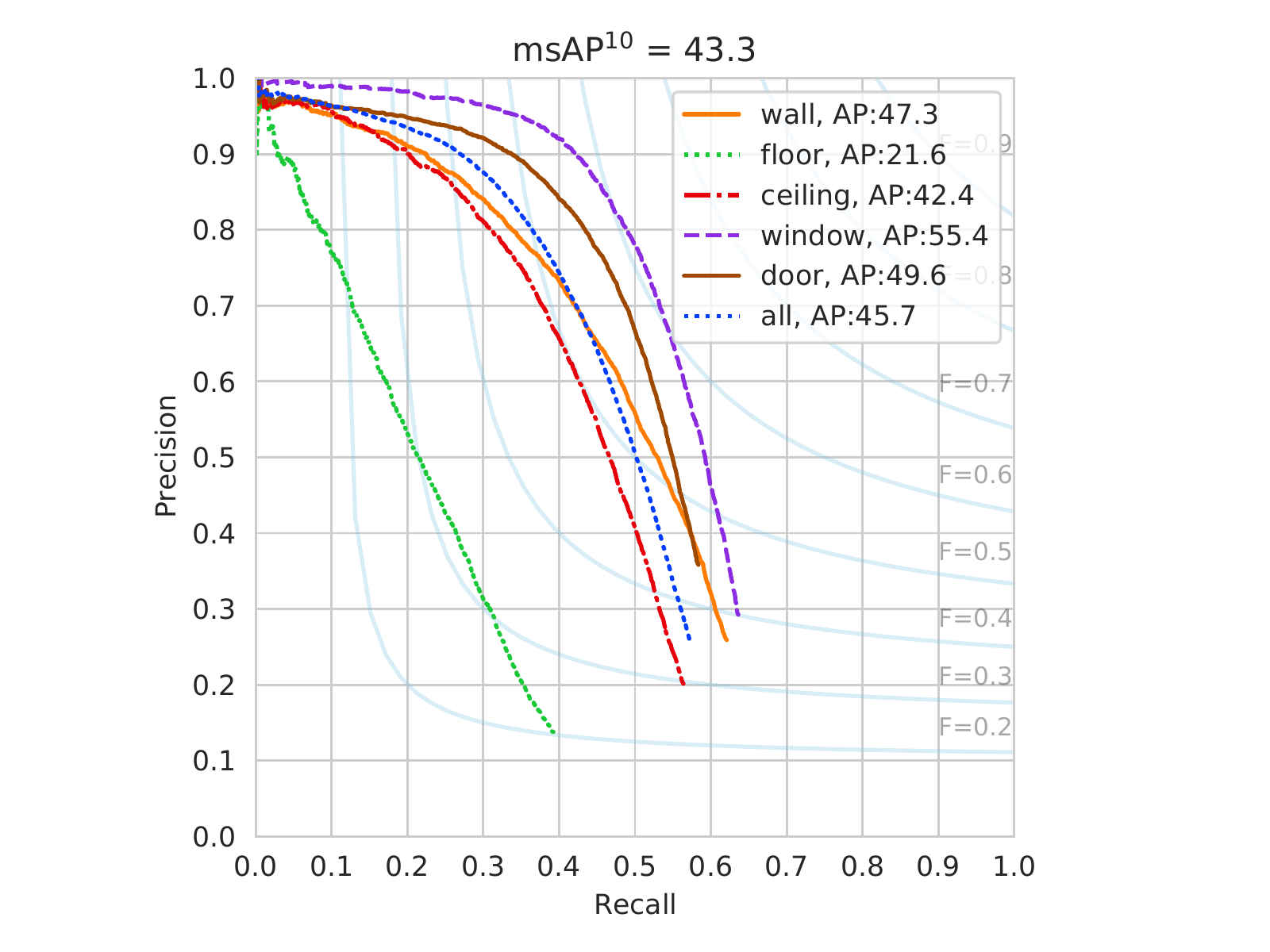}
     \hfil
     \includegraphics[trim=0 0 0 26, clip, width=\figwidth]{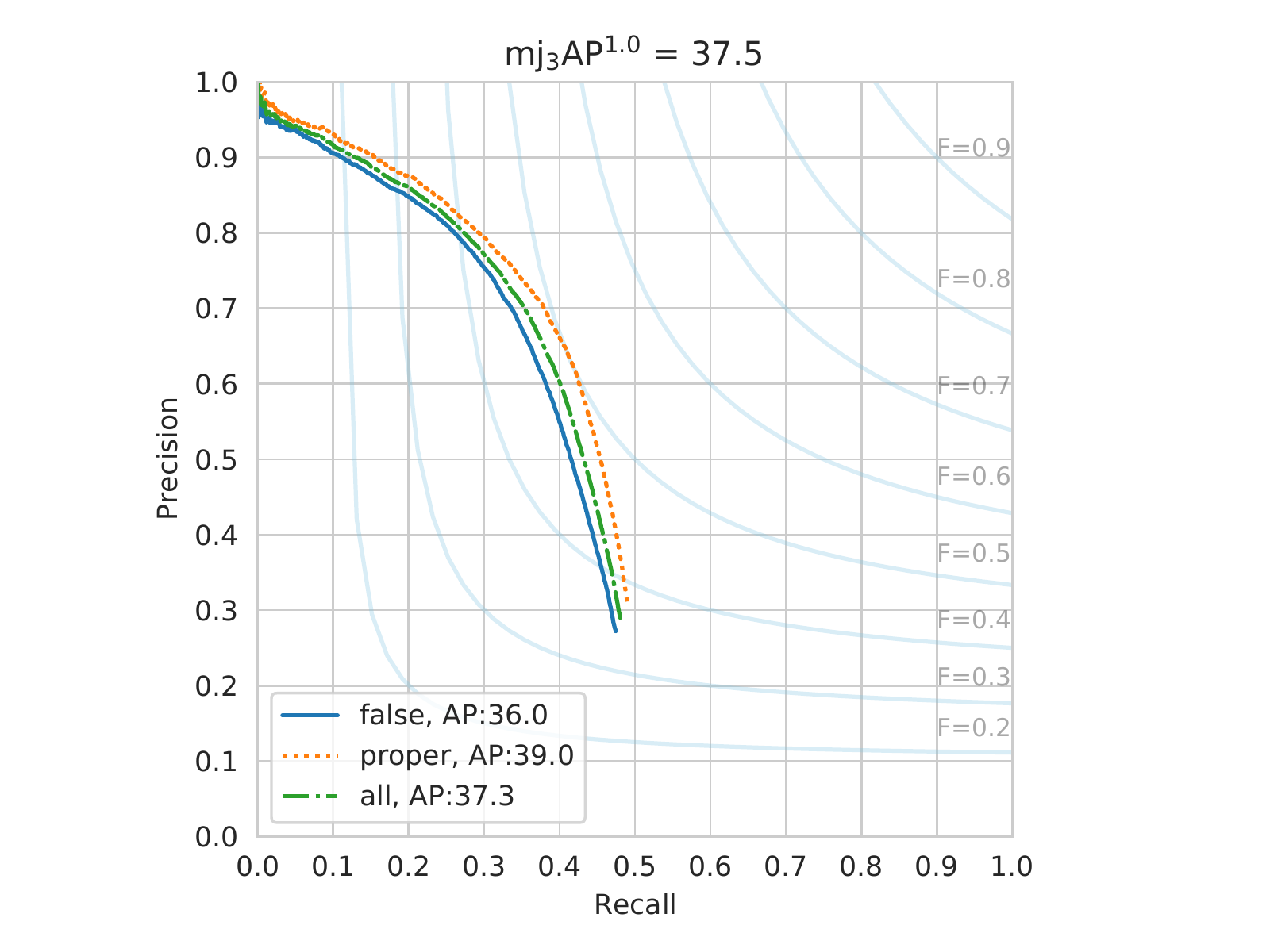}
     \caption{\textbf{Left:} $\sAP^{10}$ PR curve for line segments. \textbf{Right:} $\jAP^1$ PR curve for junctions.
     Note that \floor{} is the most difficult line label to predict.
     While all junction labels are equally difficult.}
     \label{fig:eval_PR}
\end{figure*}

\renewcommand{\figwidth}{0.195\linewidth}
\begin{figure*}[!t]
	\centering
	\includegraphics[width=\figwidth]{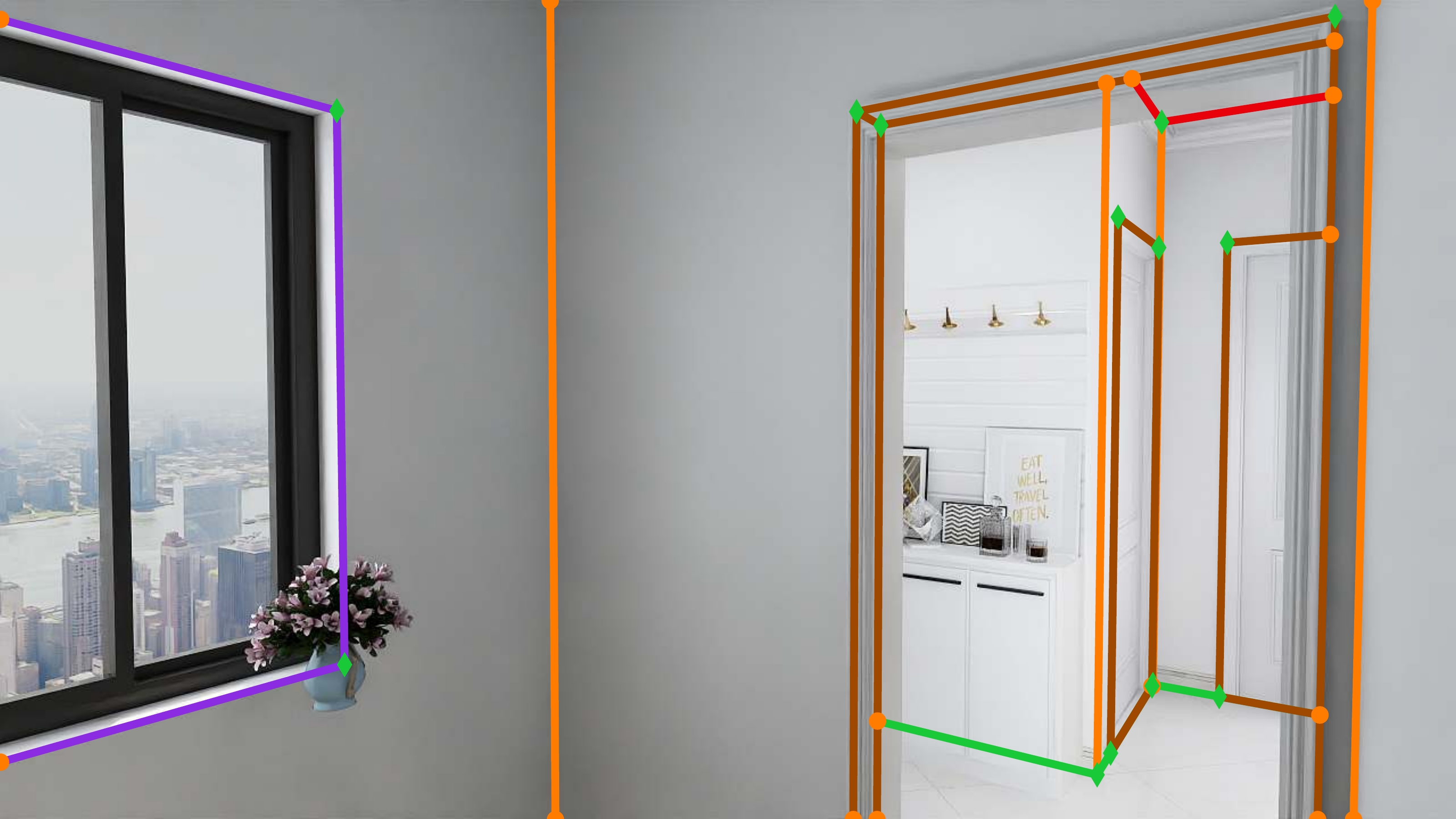}
  \includegraphics[width=\figwidth]{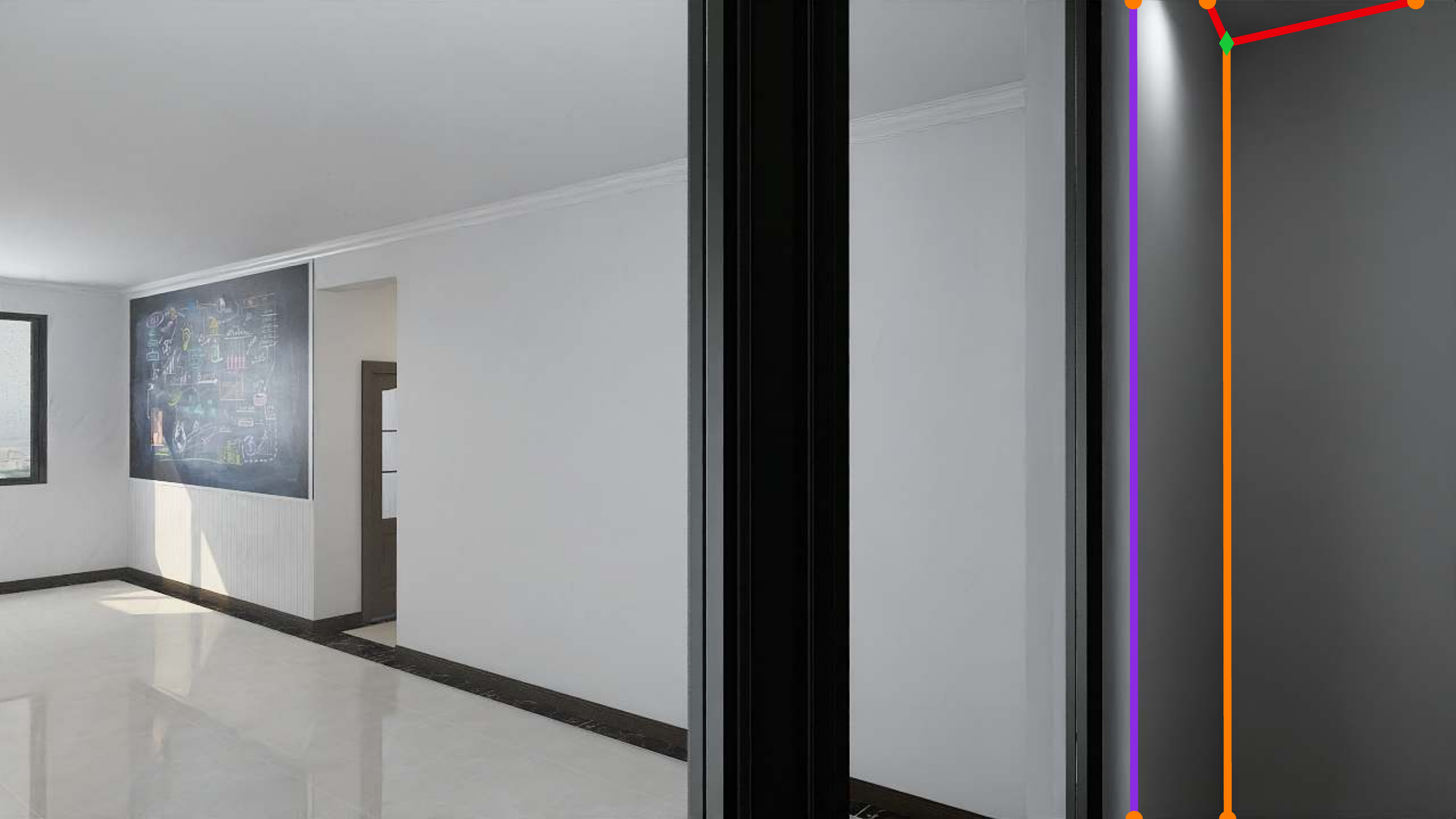}
  \includegraphics[width=\figwidth]{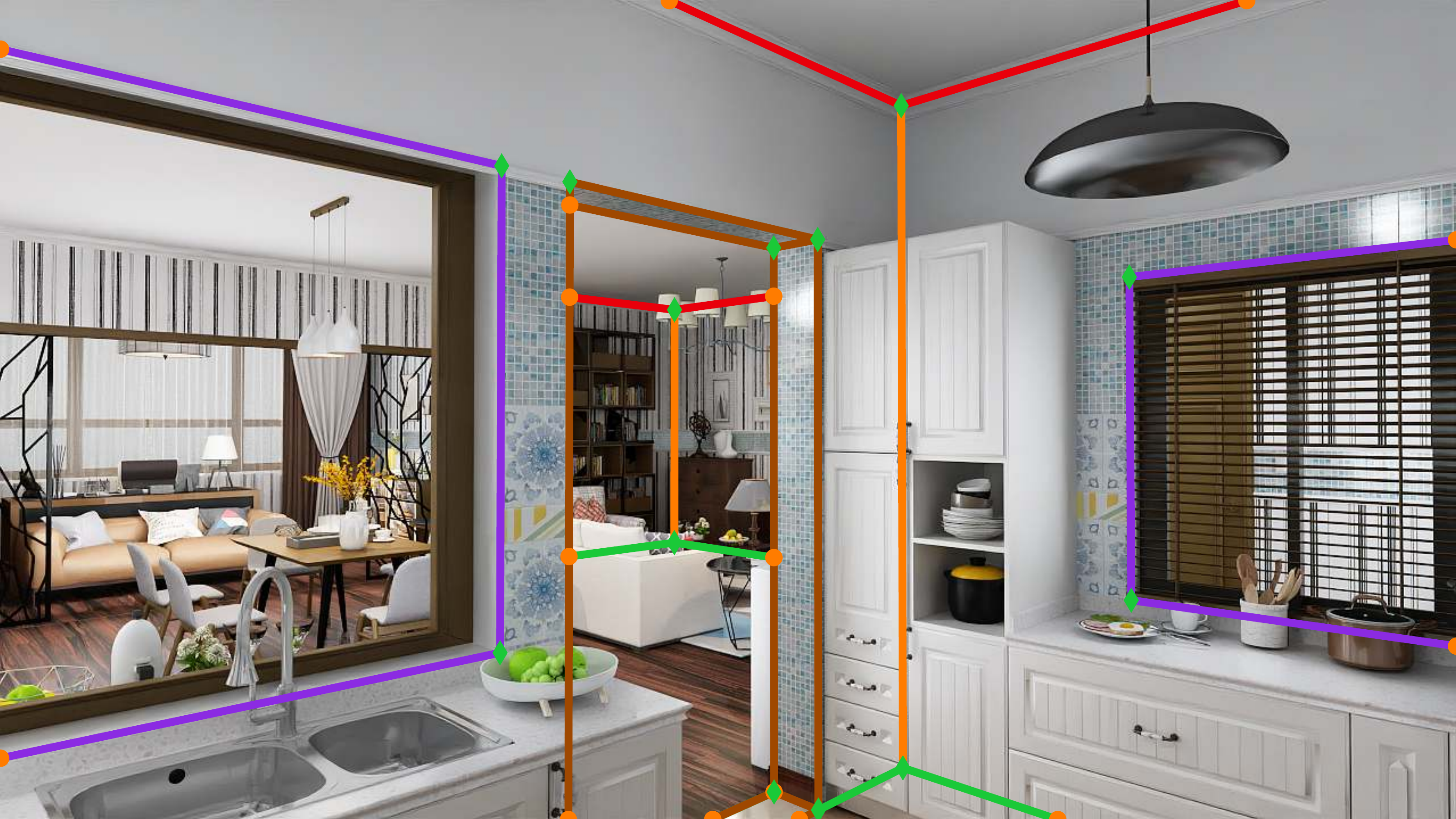}
  \includegraphics[width=\figwidth]{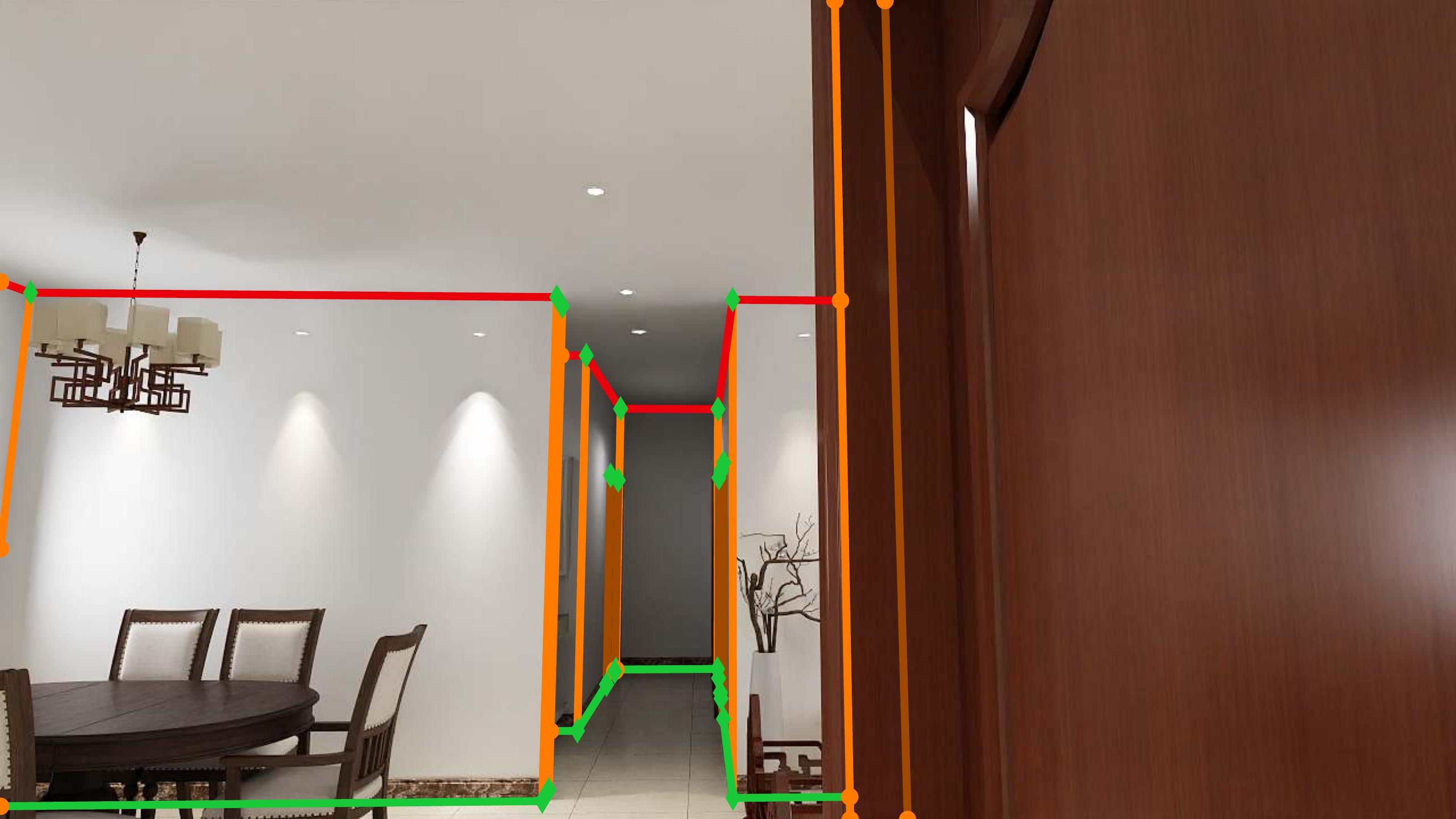}
  \includegraphics[width=\figwidth]{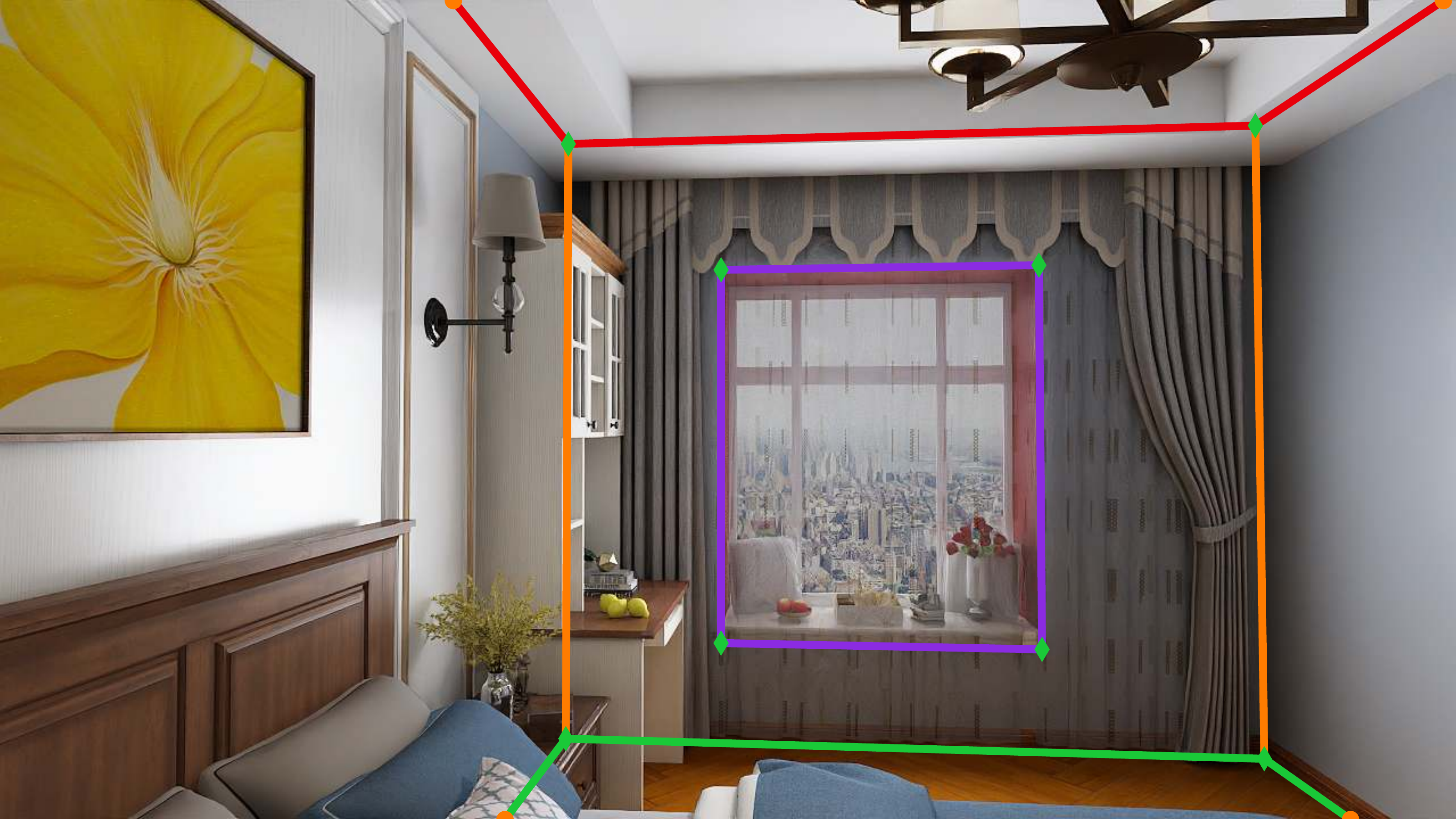}\\
  \includegraphics[width=\figwidth]{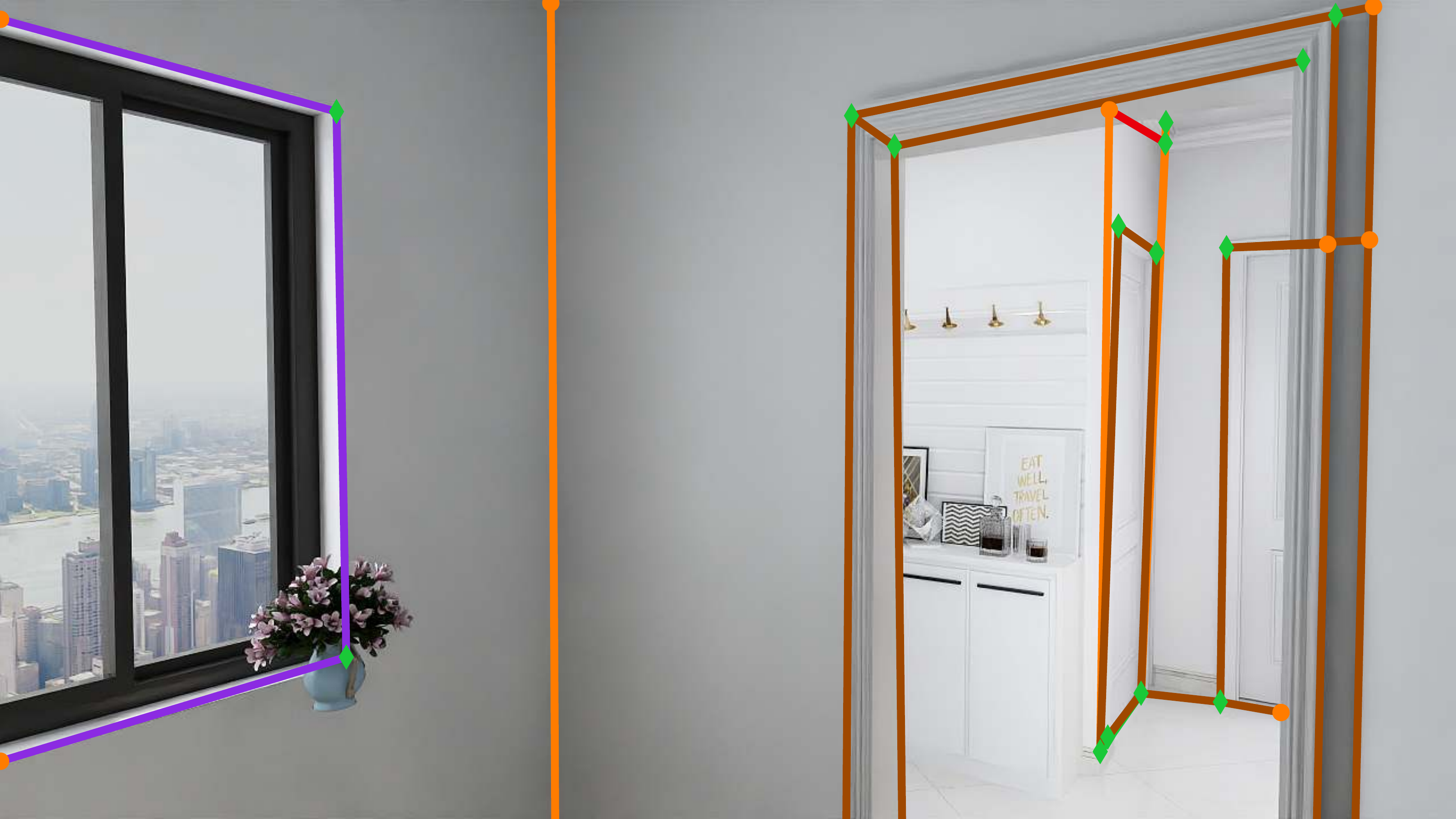}
  \includegraphics[width=\figwidth]{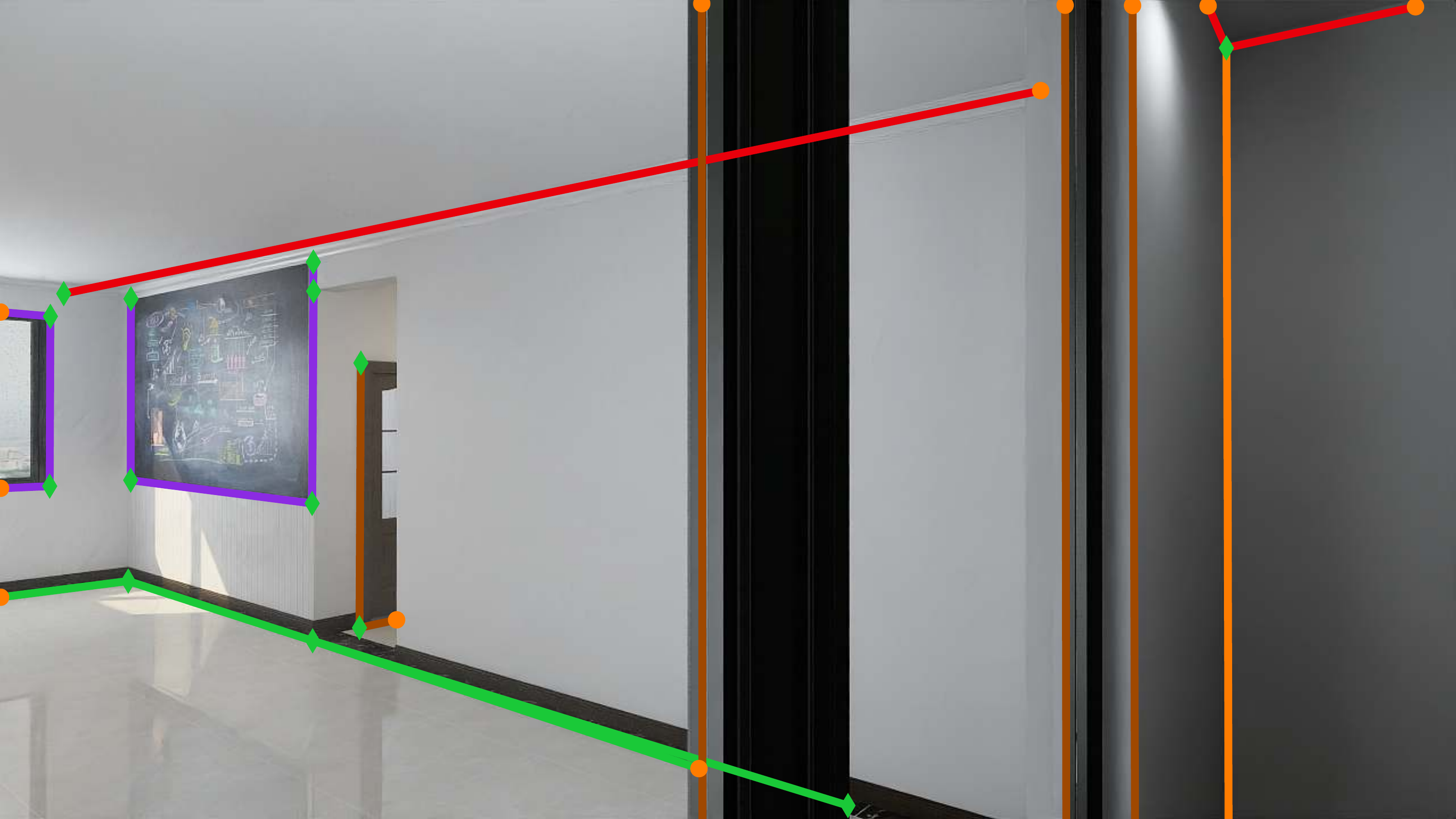}
  \includegraphics[width=\figwidth]{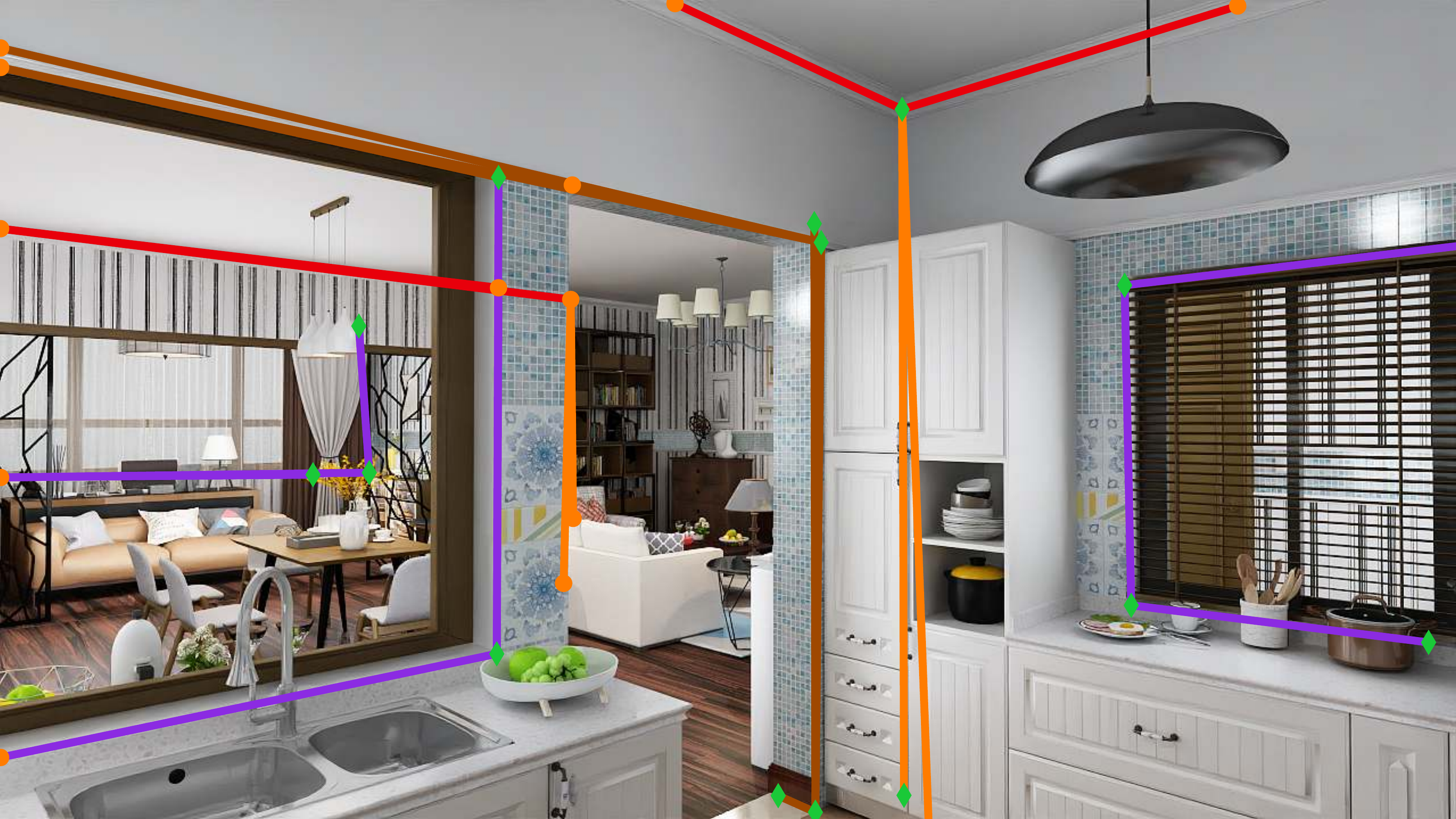}
  \includegraphics[width=\figwidth]{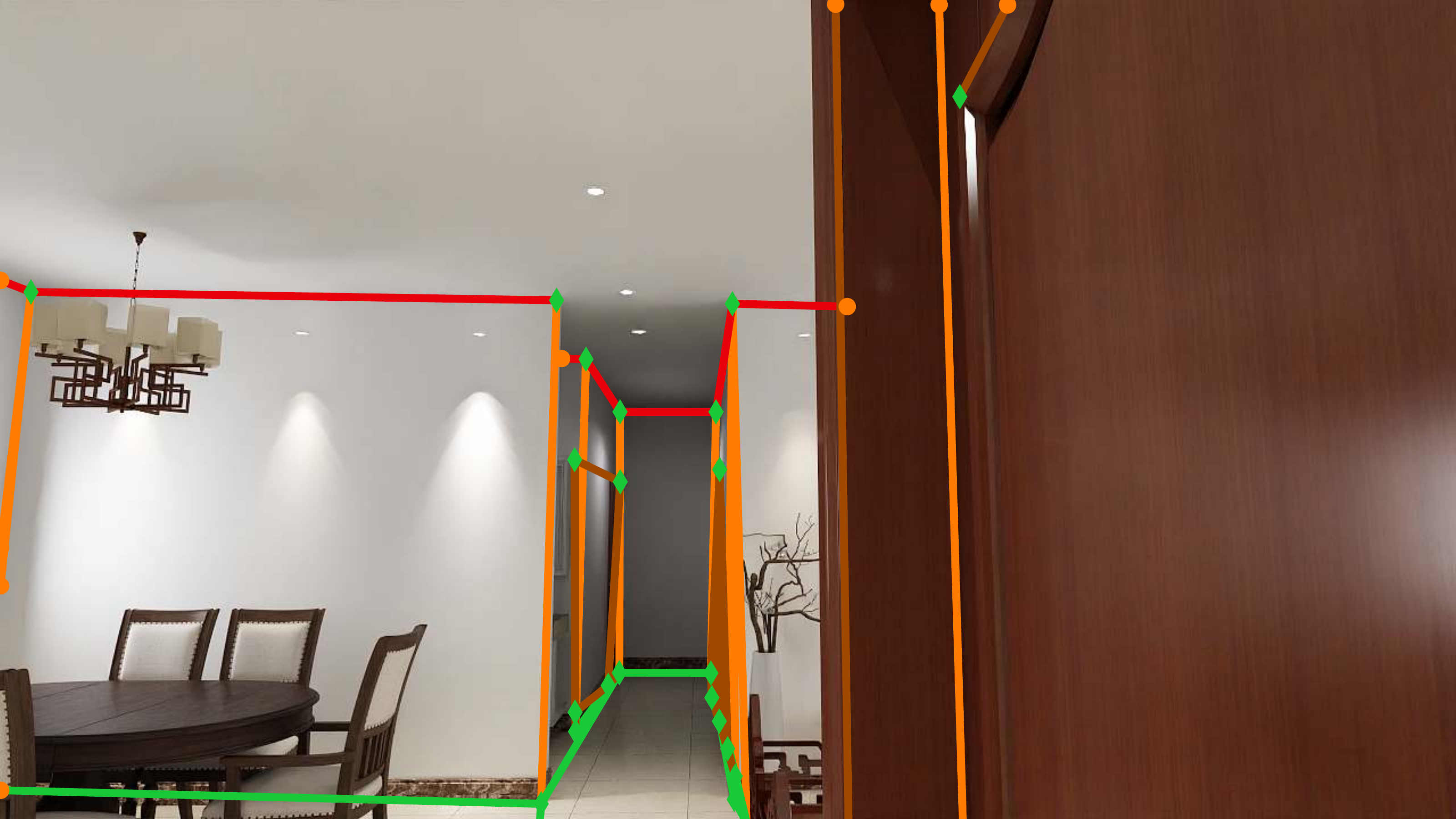}
  \includegraphics[width=\figwidth]{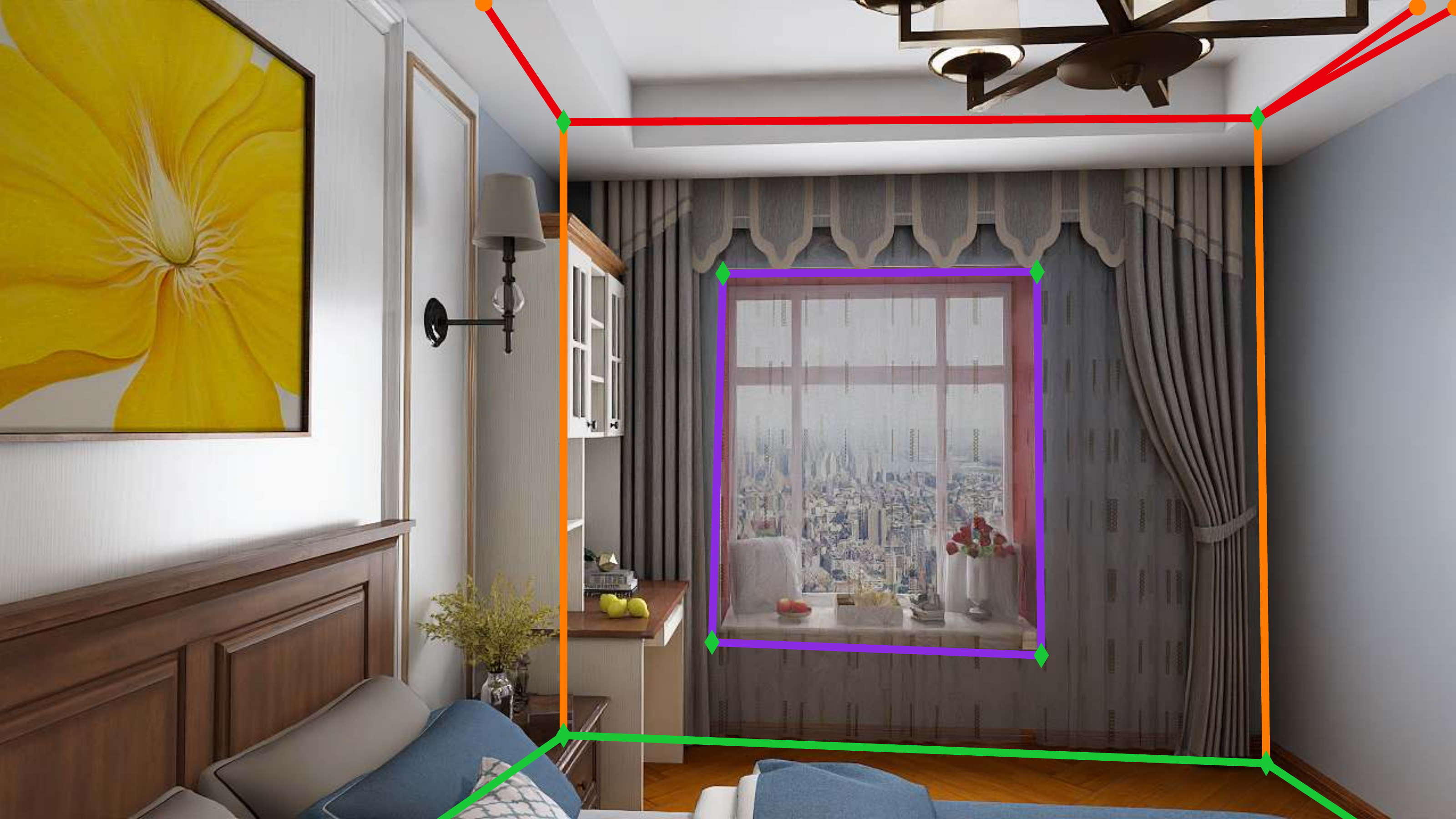}\\
  \includegraphics[width=\figwidth]{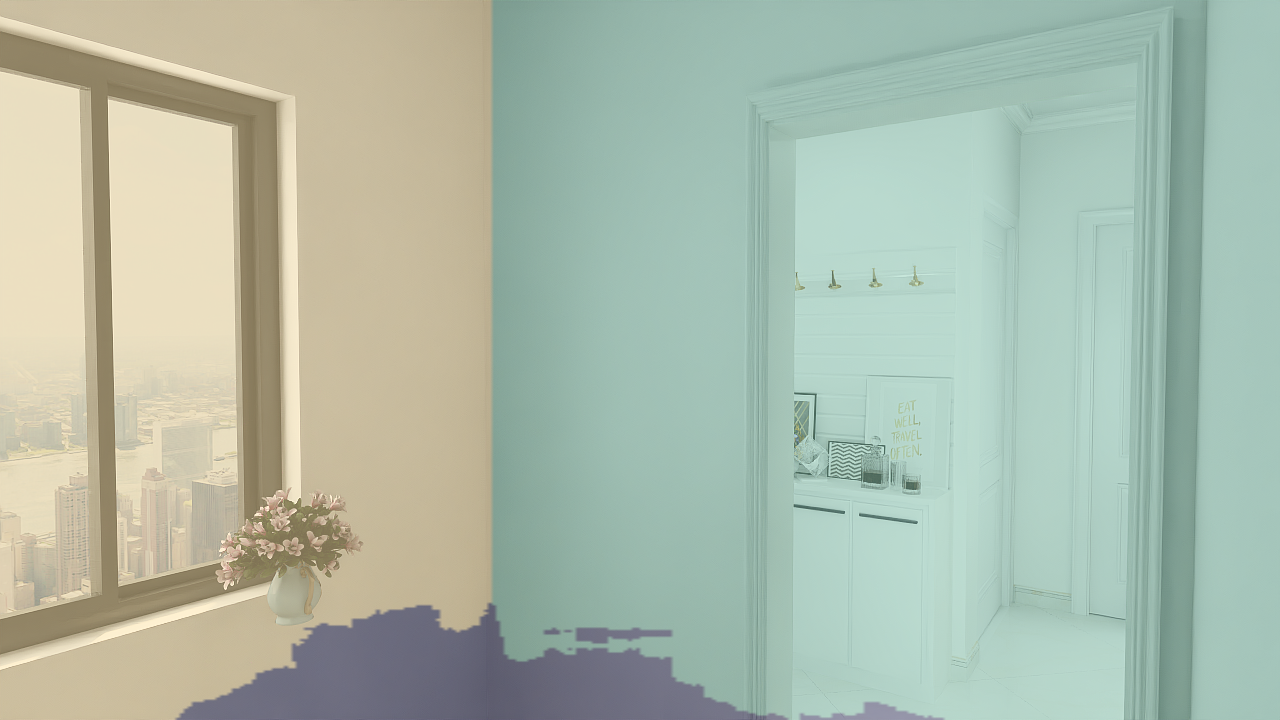}
  \includegraphics[width=\figwidth]{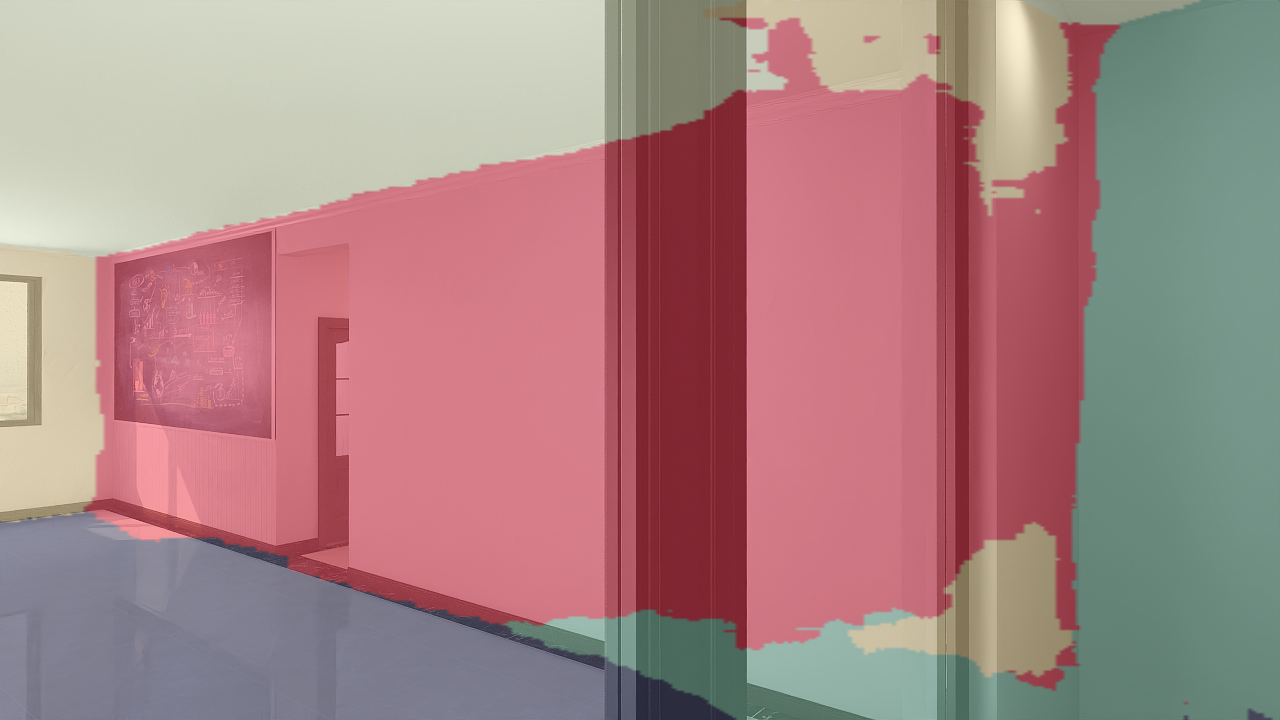}
  \includegraphics[width=\figwidth]{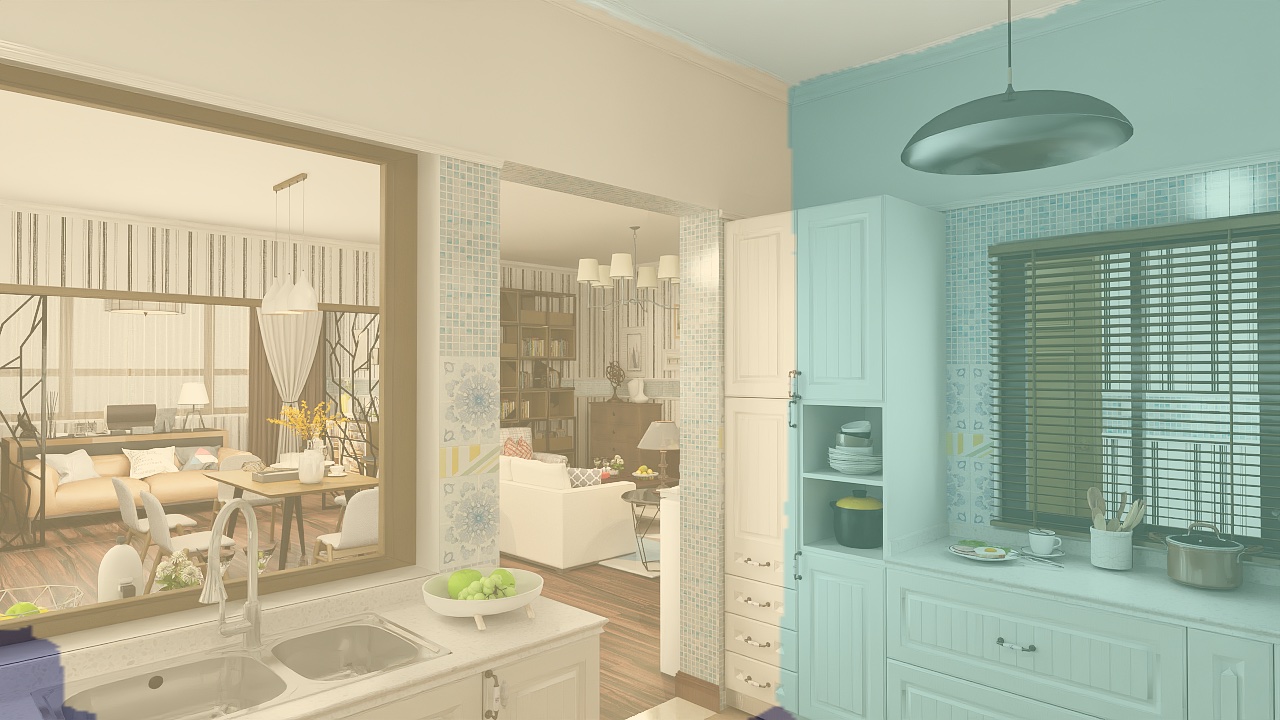}
  \includegraphics[width=\figwidth]{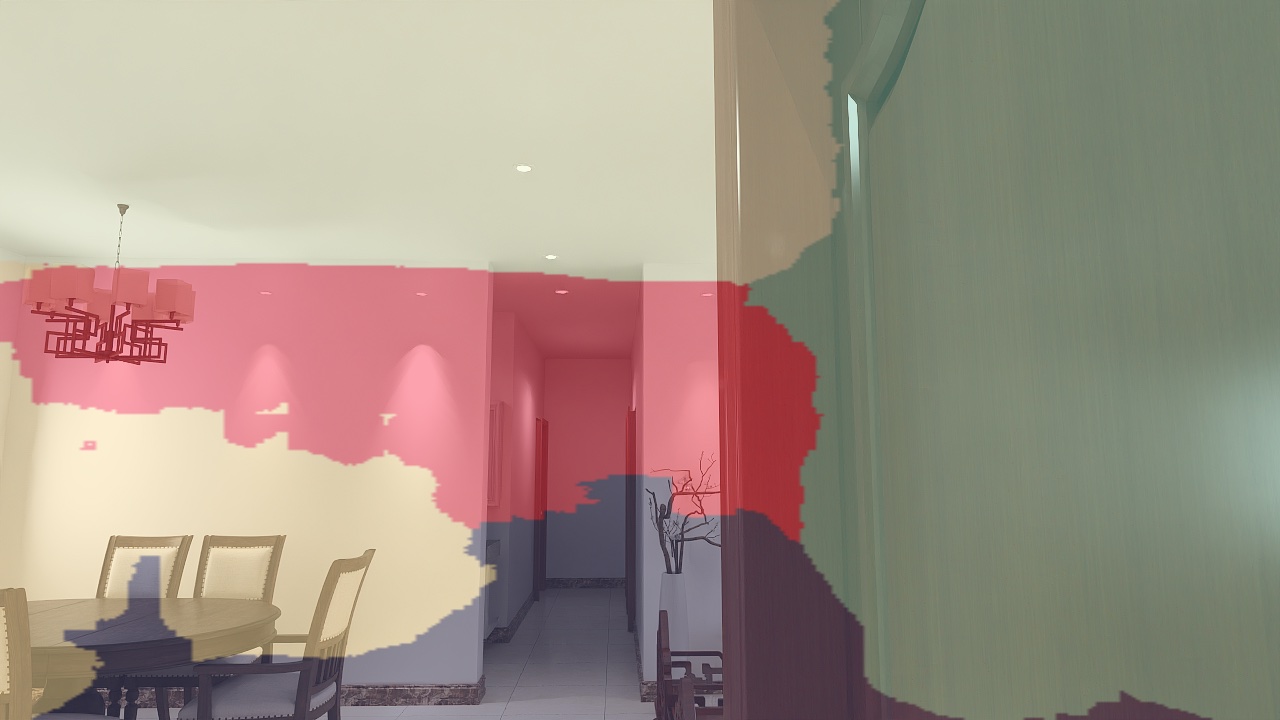}
  \includegraphics[width=\figwidth]{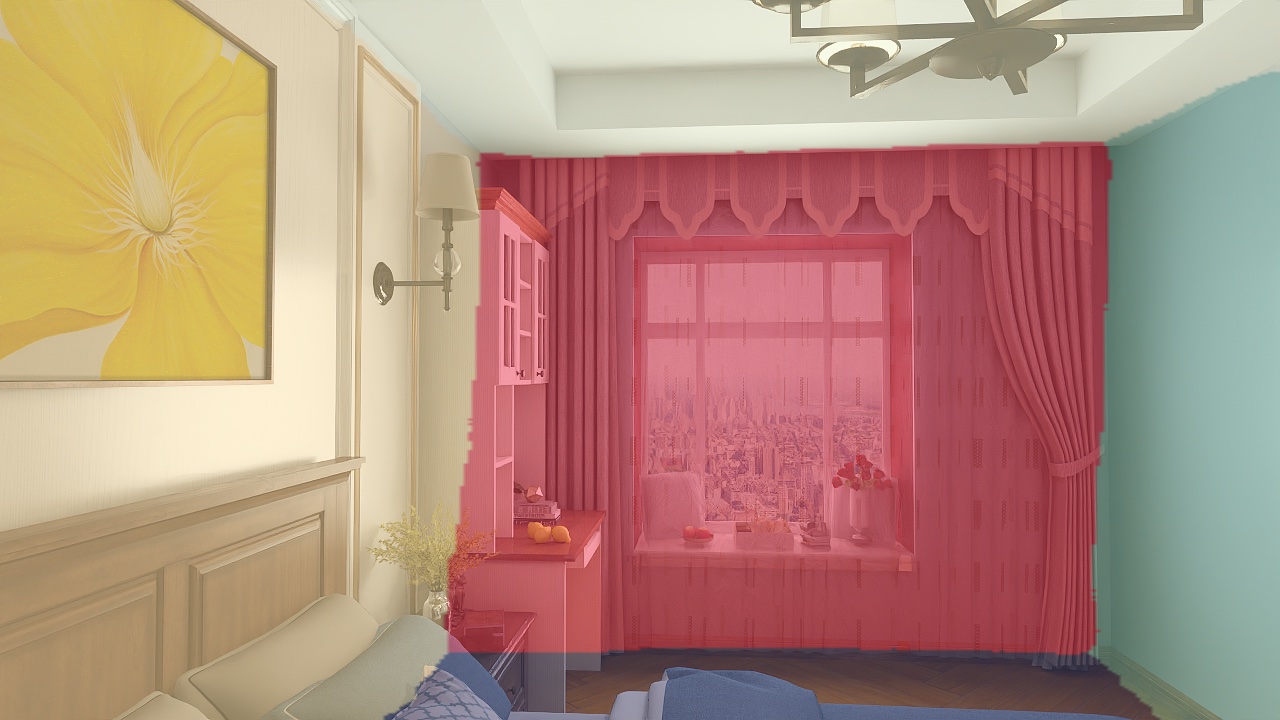}
	\caption{Results from the comparison with Room Layout Estimation. On top: the original images with Semantic Wireframe annotations. In the middle: the results from our method. In the bottom: Room Layout results from \cite{lin2018layoutestimation}.}
	\label{fig:roomlayout_vs_semwireframe}
\end{figure*}

\subsection{Ablation study}
In Table \ref{tab:gcn_ablation} we show results on how the class weights for the junction loss $\Loss(\JJ,\hat{\JJ})$,
the  GCN refinement module and line NMS affects the classification for the validation set.
Although equal weights for $\Loss(\JJ,\hat{\JJ})$ outperforms a weighted loss in  $\text{sAP}^{m}$ the
GCN refinement module benefits from the improved $\text{jAP}^{m}$ and recall.
Therefore the weighted loss with GCN refinement module and line NMS has the best performance.

\begin{table}
  \renewcommand{\arraystretch}{1.2}
  \caption{Ablation study on validation set.}
  \label{tab:gcn_ablation}
  \centering
  \begin{tabular}{|l|r|r|r|r|}
  \hline
   GCN        &  NMS         &  $\Loss(\JJ,\hat{\JJ})$  &  $\text{sAP}^{m}$  &  $\text{jAP}^{m}$ \\
  \hline
              &              &  equal                   &  44.6               & 26.4              \\
              &              &  weighted                &  31.6               & 37.4              \\
  \checkmark  &              &  equal                   &  45.0               & \textbf{39.0}      \\
  \checkmark  &              &  weighted                &  45.3               & 37.9              \\
  \checkmark  &  \checkmark  &  equal                   &  45.8               & 37.9              \\
  \checkmark  &  \checkmark  &  weighted                &  \textbf{46.7}      & 37.9              \\
  \hline
  \end{tabular}
\end{table}

\subsection{Room Layout vs Semantic Room Wireframe}
\label{sec:LSUN_vs_Structured3D}
We make a qualitative comparison in Figure \ref{fig:roomlayout_vs_semwireframe} with the method of Lin \etal \cite{lin2018layoutestimation}\footnote{\url{https://github.com/leVirve/lsun-room}} from the Room Layout Estimation task to illustrate the benefits of this data.
For our method we use all lines with score higher than 0.9.
Since the methods are trained on different data we cannot compare performance, but we see that the representation for the Room Layout Estimation task is not sufficient for these complex scenes.

\section{Conclusions}
We have introduced the task of Semantic Room Wireframe detection and generated annotations from Structured3D.
This dataset challenges algorithms to detect complex Room Geometries from single view perspective images.
We show that a CNN for Wireframe estimation can be adapted to this task and handles complex Room Geometries better than a reference method for Room Layout Estimation.
We also show how the graph structure can be used with a GCN refinement module to out-perform the baseline even in non-semantic wireframe detection.

\section*{Acknowledgment}
This work was partially supported by the strategic research projects ELLIIT and eSSENCE,
the Swedish Foundation for Strategic Research project,
Semantic Mapping and Visual Navigation for Smart Robots (grant no. RIT15-0038) and
Wallenberg Artificial Intelligence, Autonomous Systems and Software Program (WASP)
funded by Knut and Alice Wallenberg Foundation.


\IEEEtriggeratref{46}


\bibliographystyle{IEEEtran}
\bibliography{IEEEabrv,ref} 
%

\clearpage
\pagebreak

\setcounter{section}{0}
\renewcommand*{\thesection}{A-\roman{section}}
\section{Introduction}
This is supplementary material for our paper "Semantic Room Wireframe Detection from a Single View".
It further describes the data generation process and provides more images for qualitative evaluation.

\section{Structured3D-SRW Data Generation}
This section gives a brief explanation on how our dataset Structured3D-SRW, described in Section \ref{sec:dataset} of the paper, is generated.
For further details we refer to the code.

First we define for a point
$r =  \begin{pmatrix} x_1 & x_2 & \hdots & x_{n}\end{pmatrix} \in \R^n$
and its corresponding point
$p = \begin{pmatrix} y_1 & y_2 & \hdots & y_{n+1}\end{pmatrix} \in \PR^n$
in homogeneous coordinates the mapping function
\begin{equation}
\PR^n \rightarrow \R^n : \phi(p) = \begin{pmatrix} \frac{y_1}{y_{n+1}} & \frac{y_2}{y_{n+1}} & \hdots & \frac{y_n}{y_{n+1}} \end{pmatrix}.
\end{equation}
The task is to generate a set of junctions $J_i$ connected by edges $E_i$ for each image $I_i$.
To simplify we will consider a single line segment $\hat{l} = \pphat$
with endpoints $\hat{p}_1,\hat{p}_2 \in \PR^3$ s.t $\phi(\hat{p}_1),\phi(\hat{p}_2) \in \R^3$ are in the scene coordinate system.
as well as the scene's defining set of planes, which have polygons $W = \{ w_k, k=1,...,N \}$ and plane parameters
\begin{equation}
  \hat{\Pi} = \left \{ \hat{\pi}_k =
  \begin{pmatrix}
    \hat{n}_k \\ \hat{d}
  \end{pmatrix}
  \ssep \| \hat{n}_k \|=1, \phi(p) \in w_k \Rightarrow p^T \hat{\pi}_k = 0
  \right \},
\end{equation}
where $\hat{n}_k \in \R^3$ is the normal vector.

For now we assume that the line segment is in front of the camera and inside the image.
We start by transforming the planes and line segment to the camera centered coordinate system
\begin{multline}
  L = \pp = T_i \hat{L}, \quad \\
  \Pi = \left \{\pi_k =  T^{-1} \hat{\pi}_k \ssep \hat{\pi} \in \hat{\Pi} \right \}, \; \text{where} \;
T_i = \begin{pmatrix} R_i & t_i \\ 0 & 1 \end{pmatrix}.
\end{multline}
For each plane $\pi_k$ we find where on the viewing ray for line segment endpoint $p_j$ we have the plane. So if
\begin{equation}
a_j = -\frac{d_k}{\phi(p_j)^T n_k}
\end{equation}
we know that $0 < a_j < 1$ if the point is occluded. Now we have four cases
\begin{itemize}
  \item $a_1,a_2 \geq 1 \Rightarrow$ The entire line segment is in front of plane, we are done.
  \item $a_1,a_2 \leq 0 \Rightarrow$ The plane is behind the camera, we are done.
  \item $a_1,a_2 < 1 \Rightarrow$ The line segment is behind the plane, $L_b=L$.
  \item For all other cases a part of the line segment is in front of the plane.
        We form the line segment $L_b$ which is behind the plane and $L_f$ which is in front.
\end{itemize}
Now we can project $L_b$ onto the plane $\pi_k$ and take the 2D geometric difference to get a set of line segments
\begin{equation}
  \hat{L}_w = \left ( \begin{pmatrix} a_1 \\ a_2 \end{pmatrix} \odot L_b \right) - w_k,
\end{equation}
where $\odot$ is elementwise multiplication. These are in the plane $\pi_k$ so we so we find the corresponding line segments on $L$ and denote them $L_w$.
We then merge $L_f$ with $L_w$ and get the visible set of line segments which we transform back to the camera coordinate system and apply $K_i$ to get the pixel coordinates.
An overview of the algorithm is given in Algorithm \ref{alg:annotate}. Please refer to the implementation code for details.

\begin{algorithm}
\SetKwFunction{makePolygons}{makePolygons}
\SetKwFunction{inFront}{cutInFrontAndInImage}
\SetKwFunction{behindPlane}{behindPlane}
\SetKwFunction{appendLine}{appendLine}
\SetKwFunction{inPolygon}{inPolygon}
\SetKwFunction{continue}{continue}
\SetAlgoLined
\KwData{
Parameters of scene planes: $\hat{\Pi}$,
Semantic images: $\hat{\textbf{I}}$,
RGB images: $\textbf{I}$,
All plane junctions: $\textbf{J}$,
All scene lines: $\hat{\textbf{L}}$,
}
\KwResult{Set of visible line segments $\textbf{L}^i_V$ for each image $i$}
 $W :=$ \makePolygons{$\hat{\textbf{I}}, \hat{\Pi},\textbf{J}$}\;
 \For{$I_i \in \textbf{I}$}{
   $\Pi := T^{-1}_i \hat{P_i}$\;
   $\textbf{L} := T_i \hat{\textbf{L}}$\;
   \For{$L \in \textbf{L}$}{
    $L:= $\inFront{$L$}\;
    \If{$L = \emptyset$}{
      \continue\;
    }
    $\textbf{L}^i_V = \{ L \}$ \;
    \For{$w_k \in W, \pi_k \in \Pi$}{
      $L_t := \emptyset$\;
      \For{$L \in L^i_V$}{
        $L_f, L_b := $ \behindPlane{$\pi_k,L$}\;
        $\textbf{L}_p := $ \inPolygon{$L_b$}\;
        $L_t := L_t \cup$ \appendLine{$\textbf{L}_p, L_f$}\;
      }
      $\textbf{L}^i_V := L_t$\;
    }
   }
  }
 \caption{An overview of how the algorithm to annotate one scene works. For details and special cases please refer to the implementation code.}
 \label{alg:annotate}
\end{algorithm}

\section{Plane estimation}
\begin{figure}
  \centering
  \includegraphics[trim=0 165 0 30, clip, width=\linewidth]{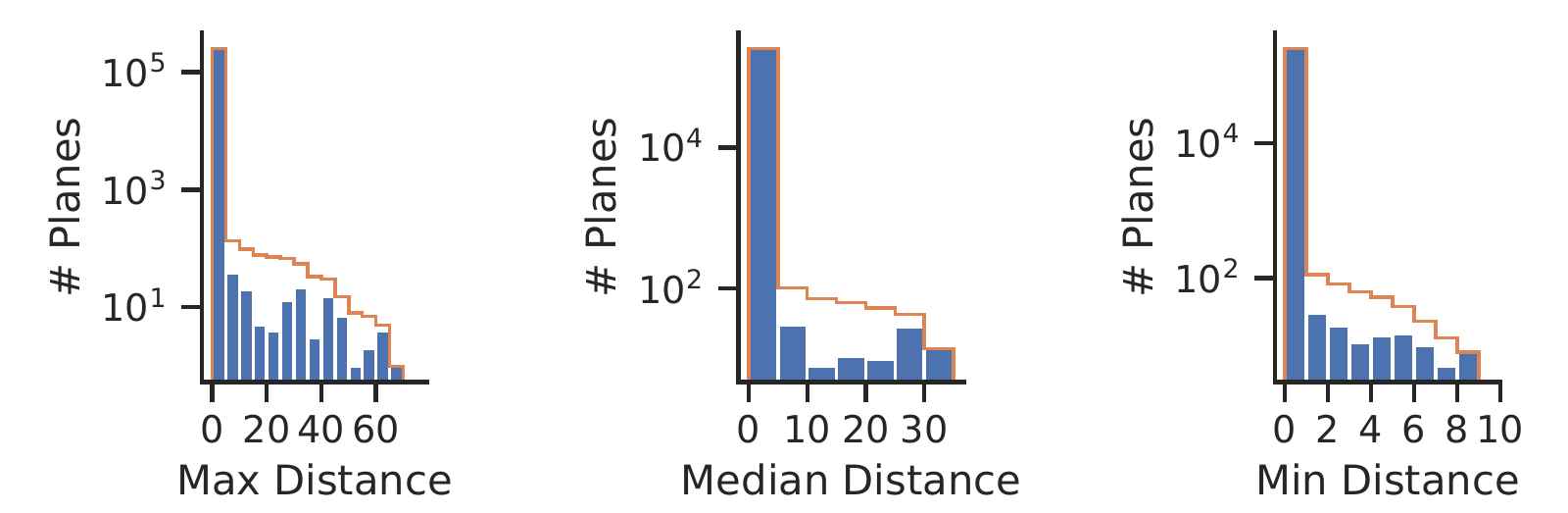}%
  \caption{Histograms for the maximum, median and minimum distance from each plane to its junctions using the estimated plane equation.}
  \label{fig:wall_hist_est}
\end{figure}

In Section \ref{sec:analysis} we estimate new plane parameters that fit optimally with the plane junction.
See Figure \ref{fig:wall_hist_est} for the histograms of distances from plane to junctions.

\section{Additional Images for Qualitative Comparison}
Here is and extension to Section \ref{sec:LSUN_vs_Structured3D} of the paper, where we compared SRW-Net a Room Layout estimation algorithm \cite{lin2018layoutestimation}.
For additional images from Structured3D-SRW, see Figure \ref{fig:roomlayout_vs_semwireframe2}.

\newpage

\renewcommand{\figwidth}{0.32\linewidth}
\begin{figure*}[!t]
	\centering

  \includegraphics[width=\figwidth]{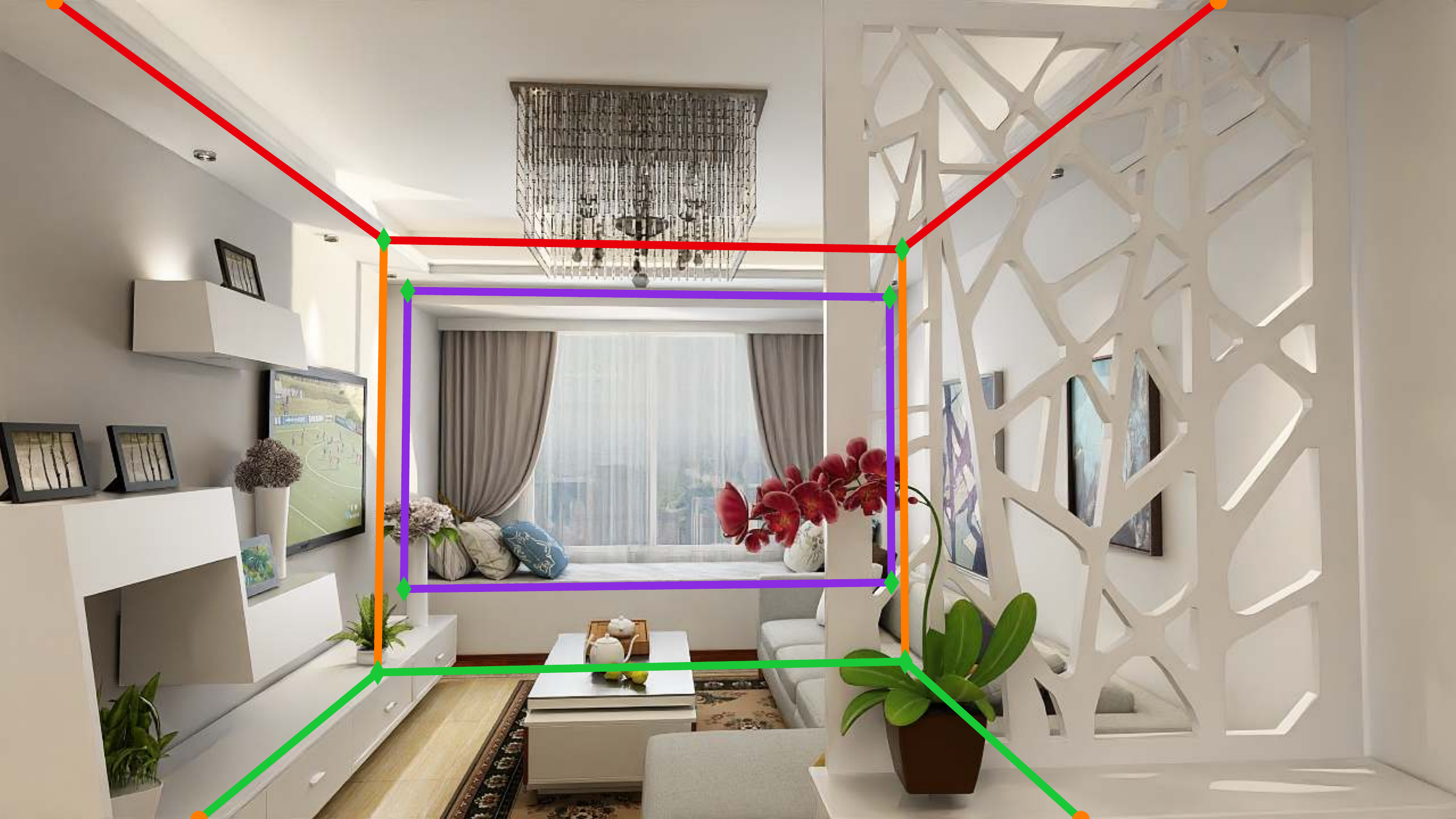}
  \includegraphics[width=\figwidth]{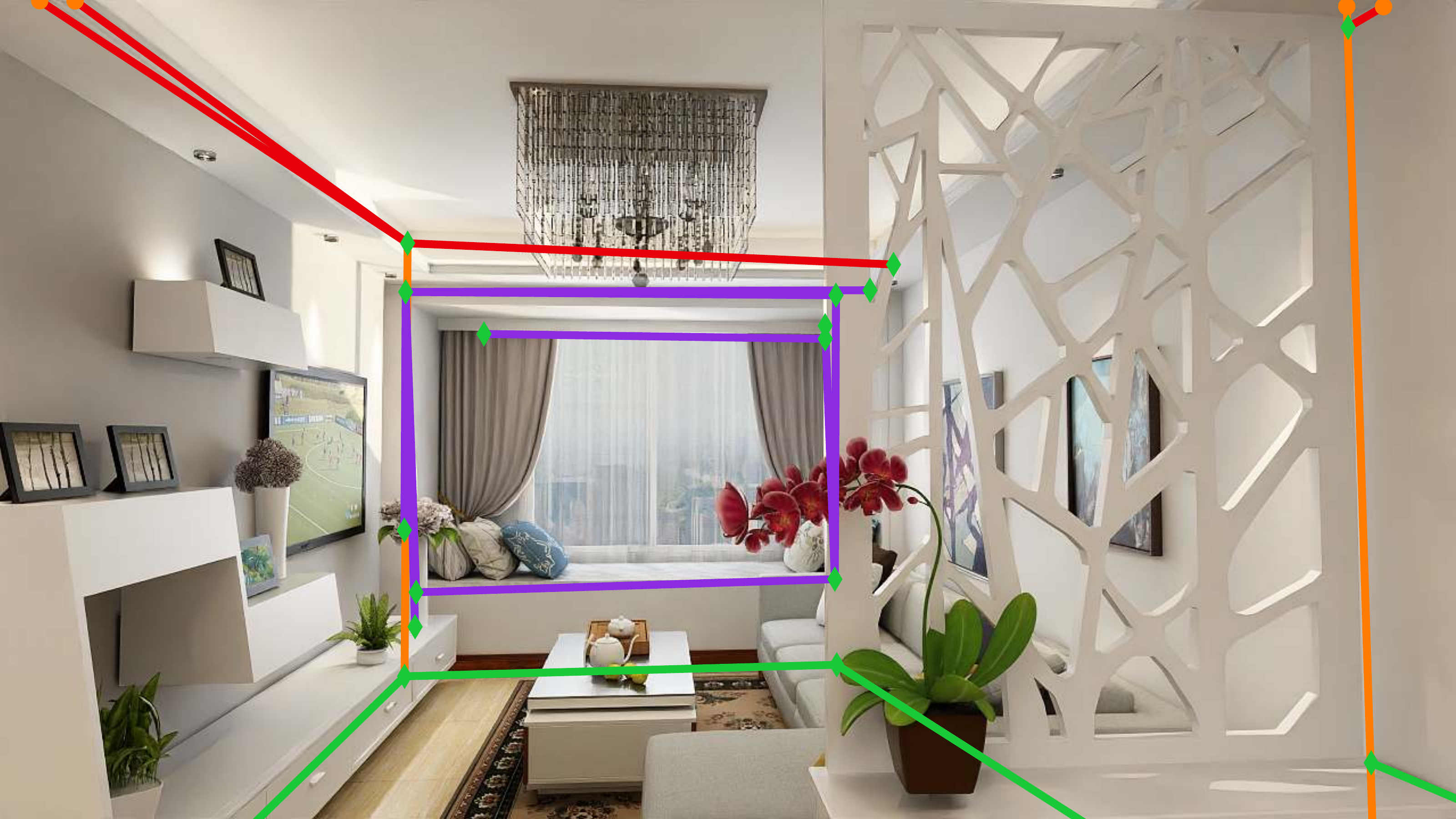}
  \includegraphics[width=\figwidth]{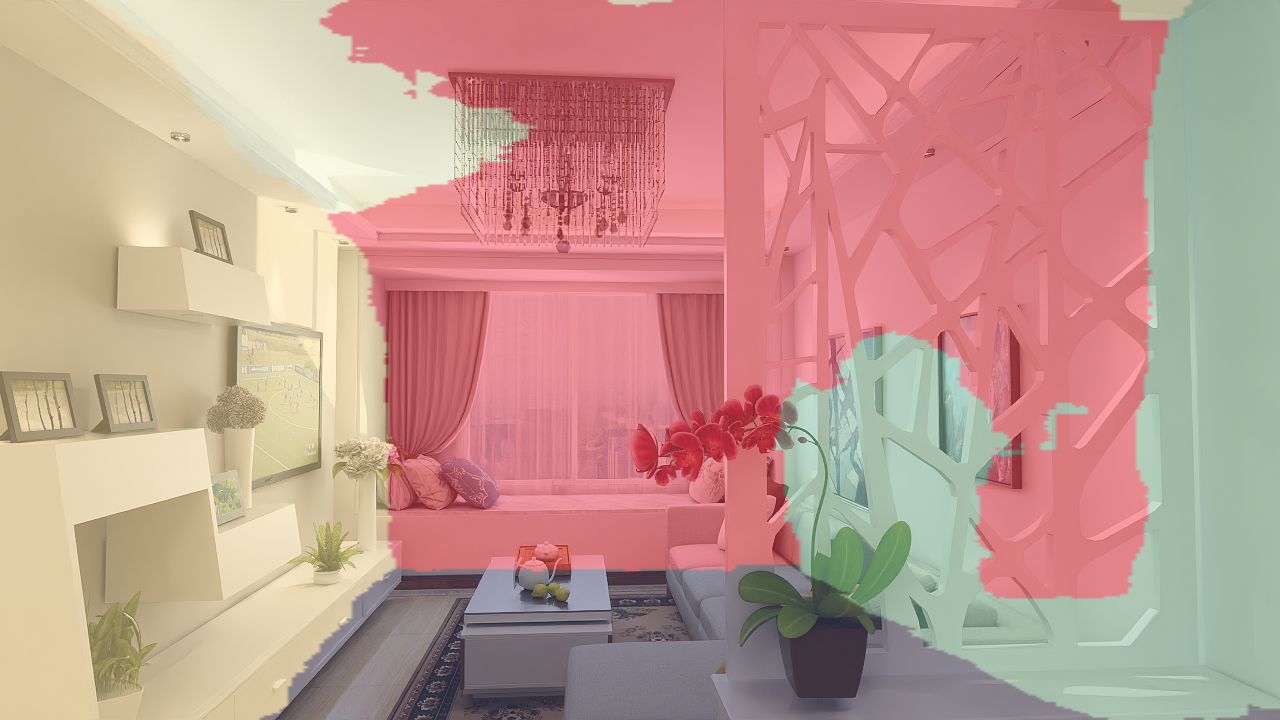}\\

  \includegraphics[width=\figwidth]{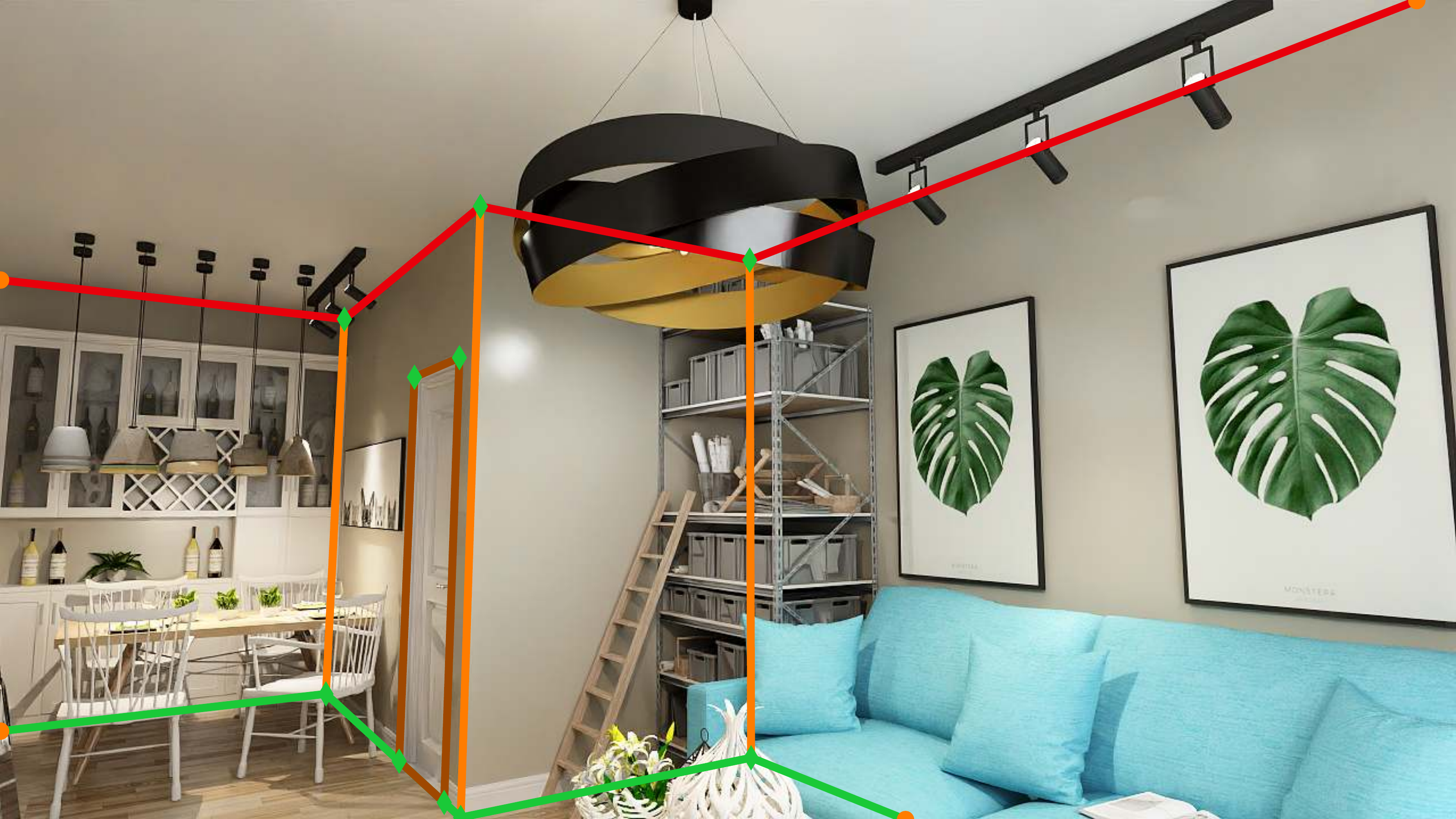}
  \includegraphics[width=\figwidth]{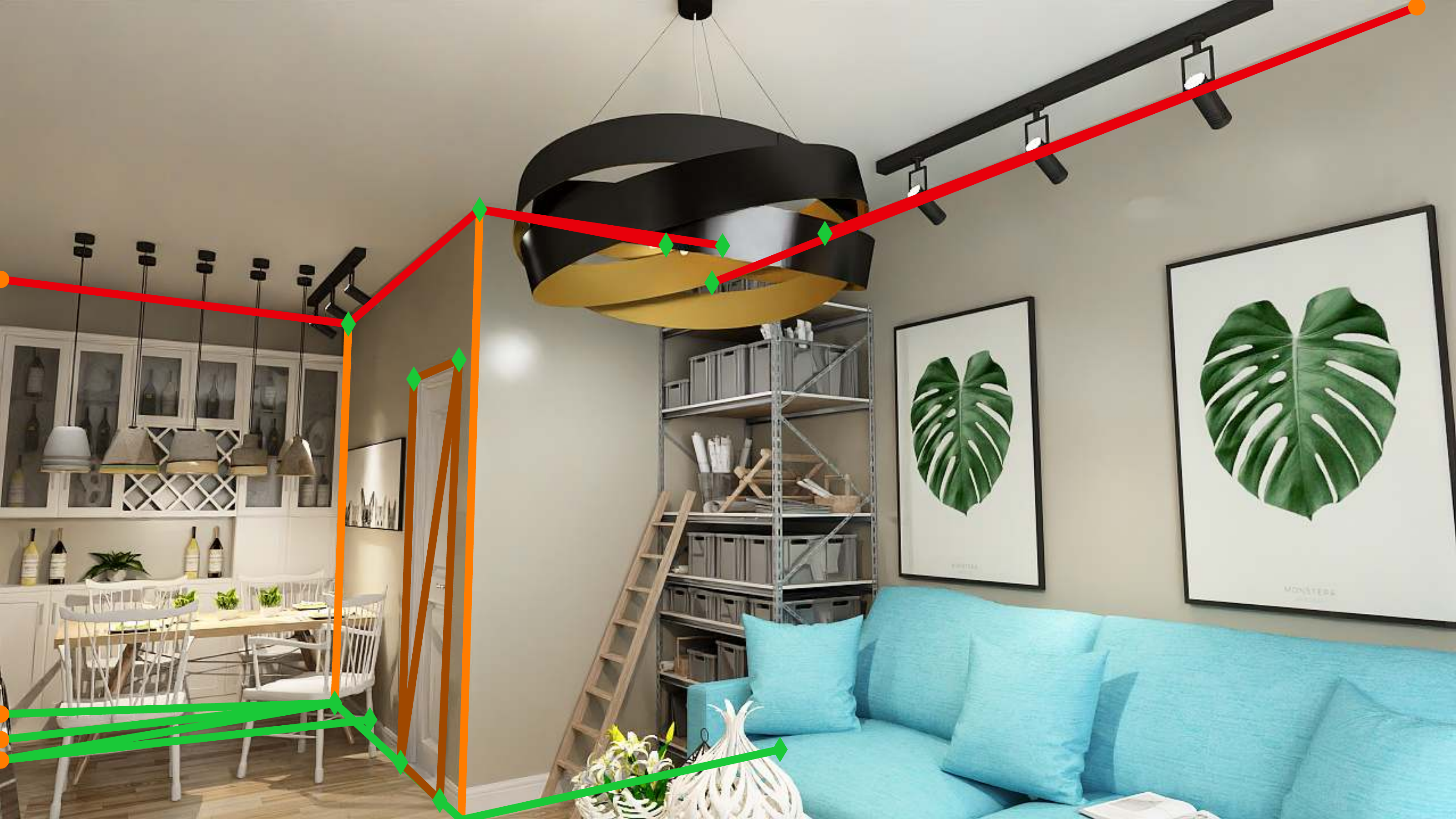}
  \includegraphics[width=\figwidth]{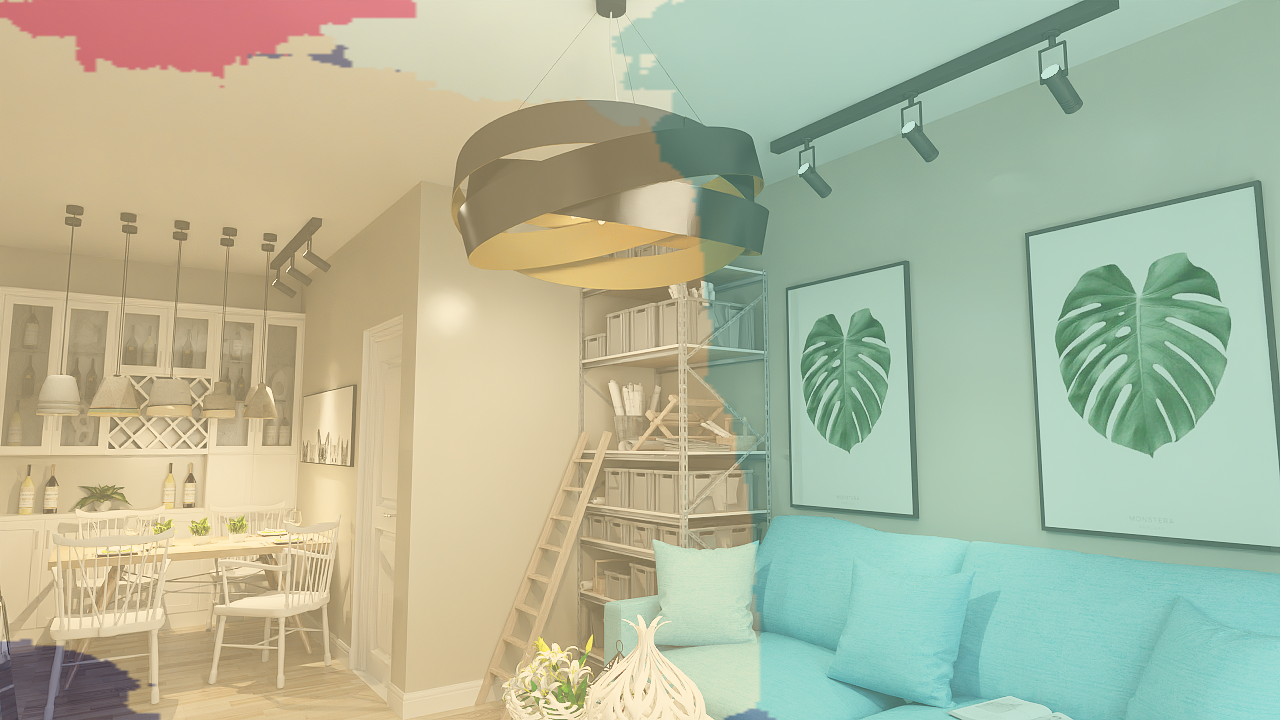}\\

  \includegraphics[width=\figwidth]{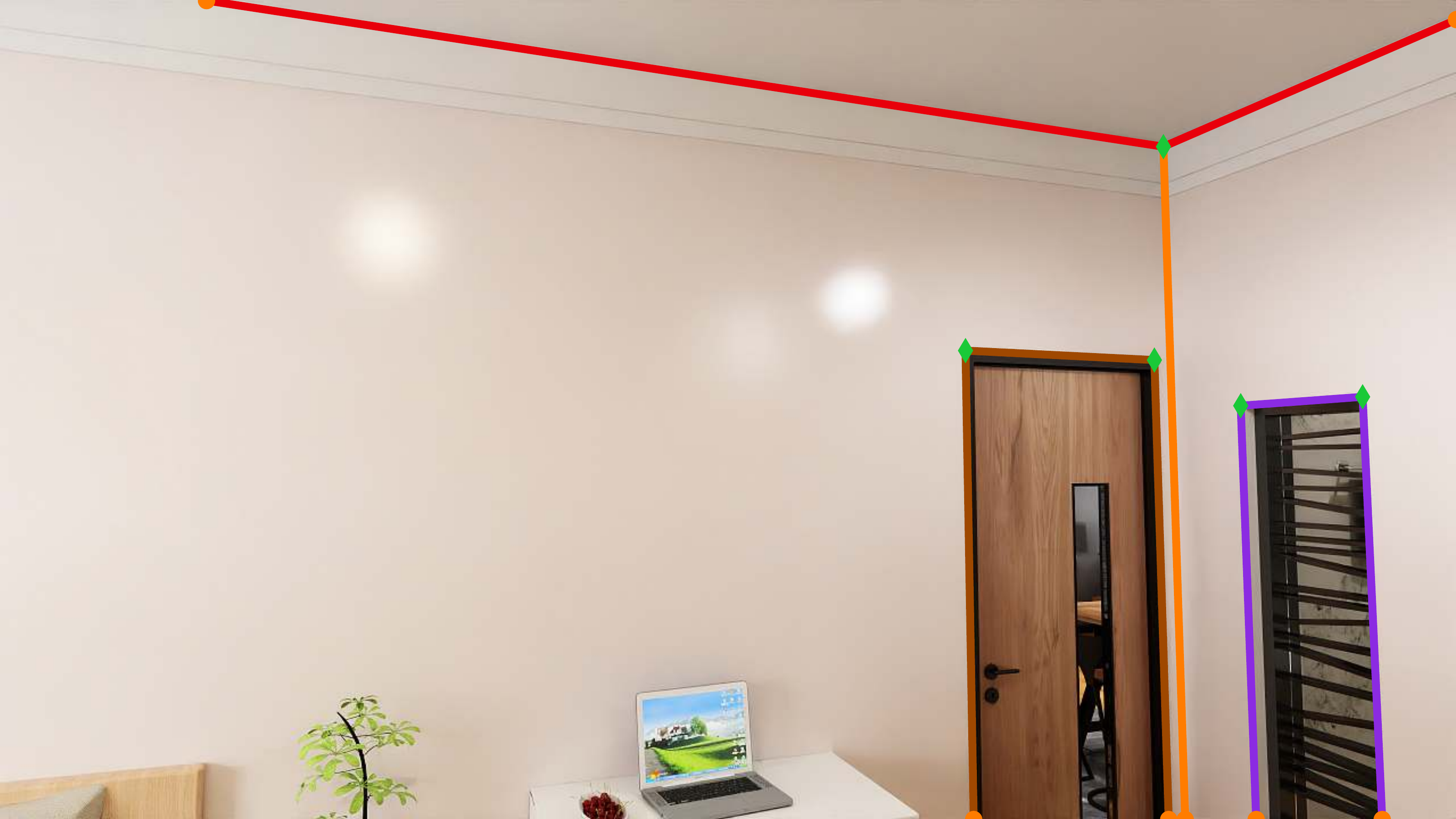}
  \includegraphics[width=\figwidth]{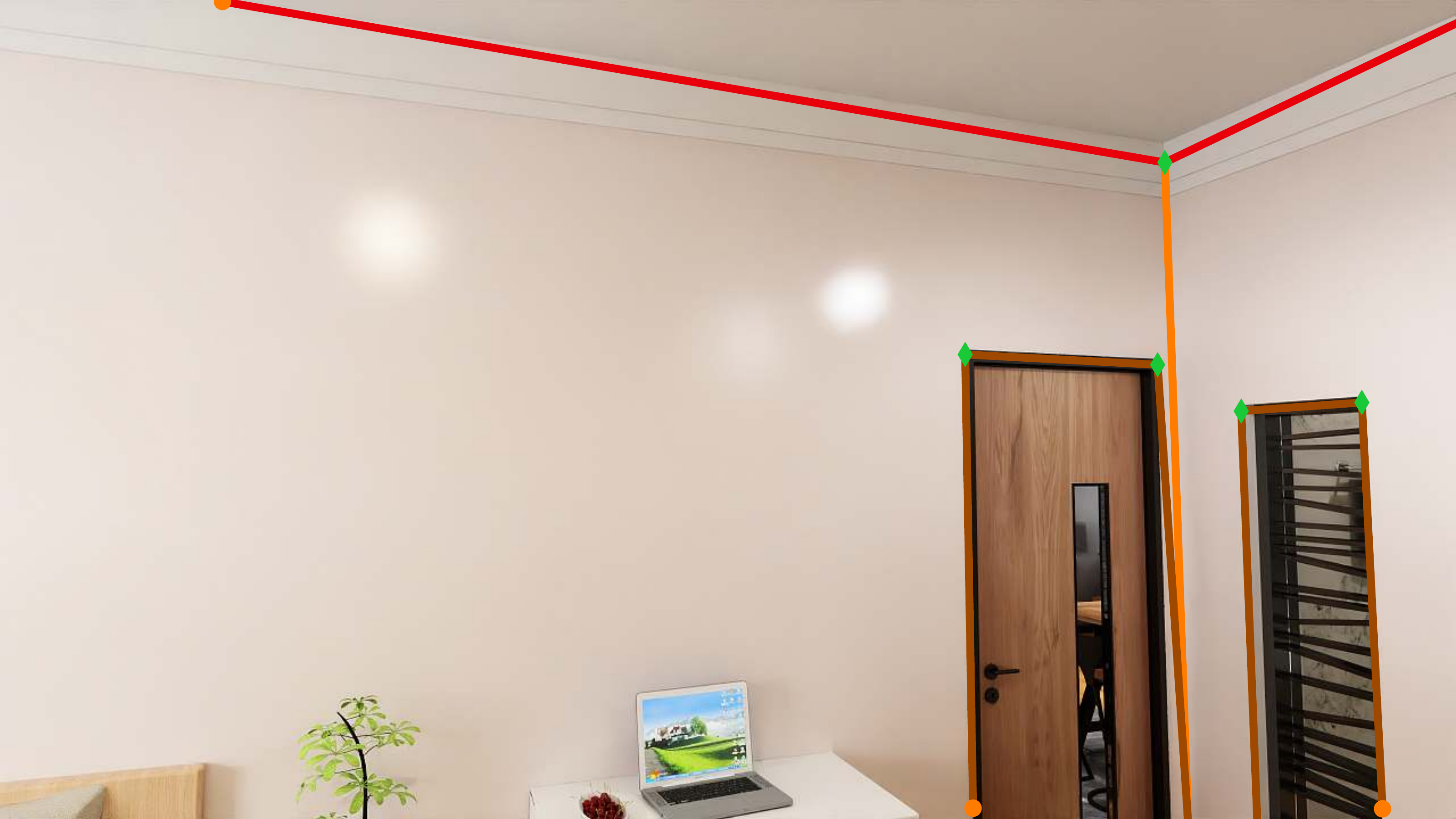}
  \includegraphics[width=\figwidth]{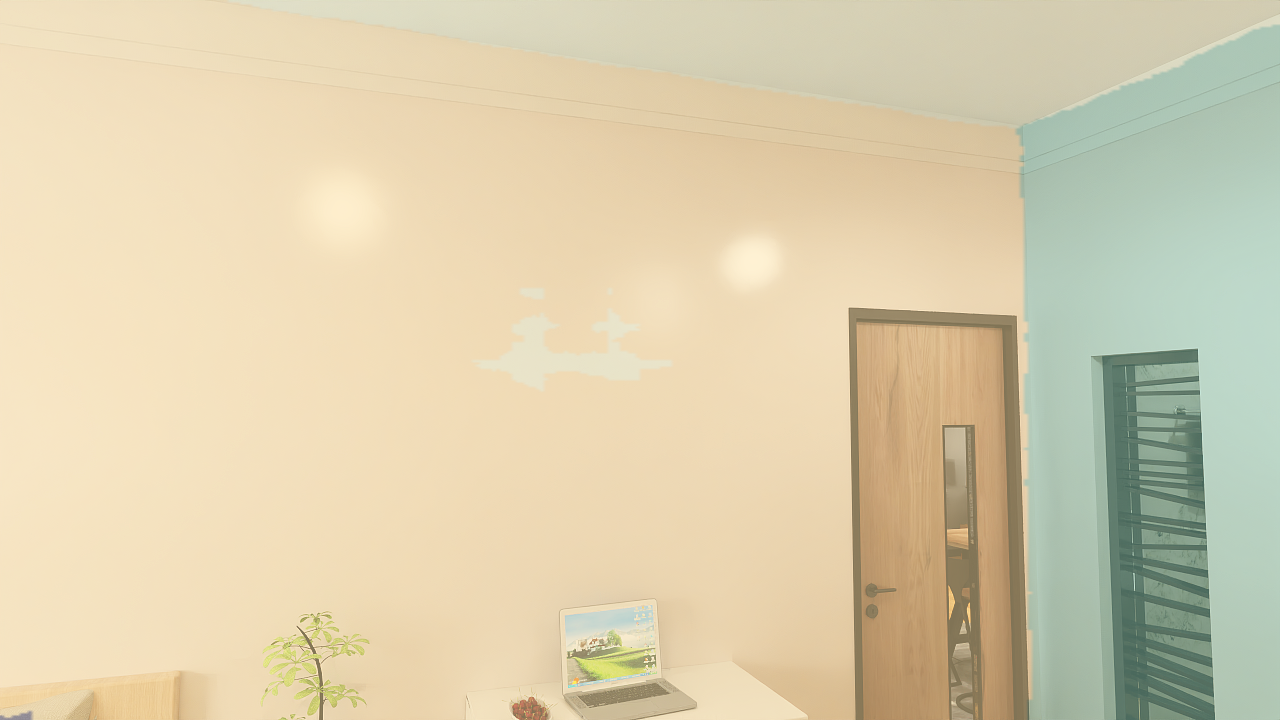}\\

  \includegraphics[width=\figwidth]{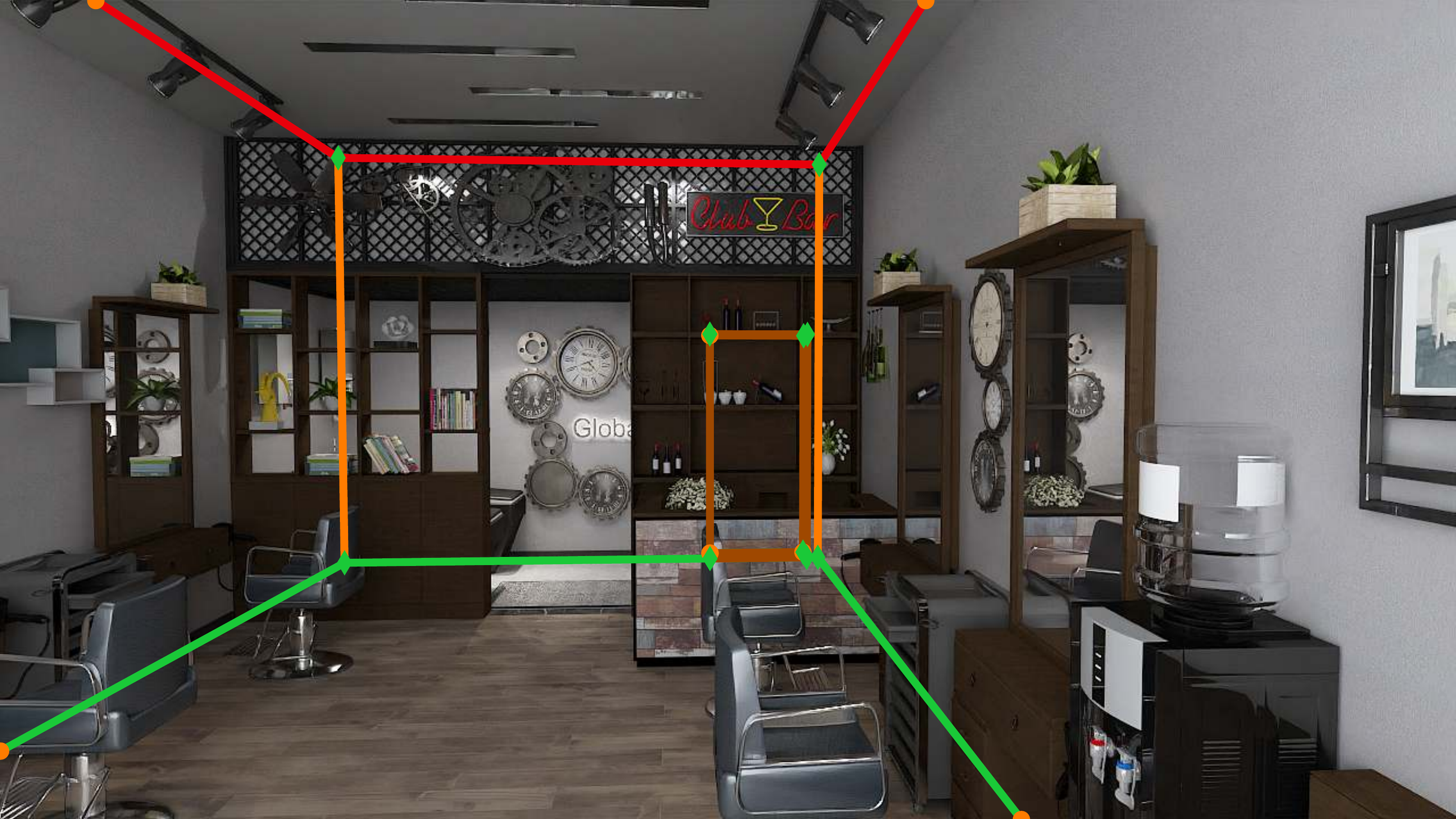}
  \includegraphics[width=\figwidth]{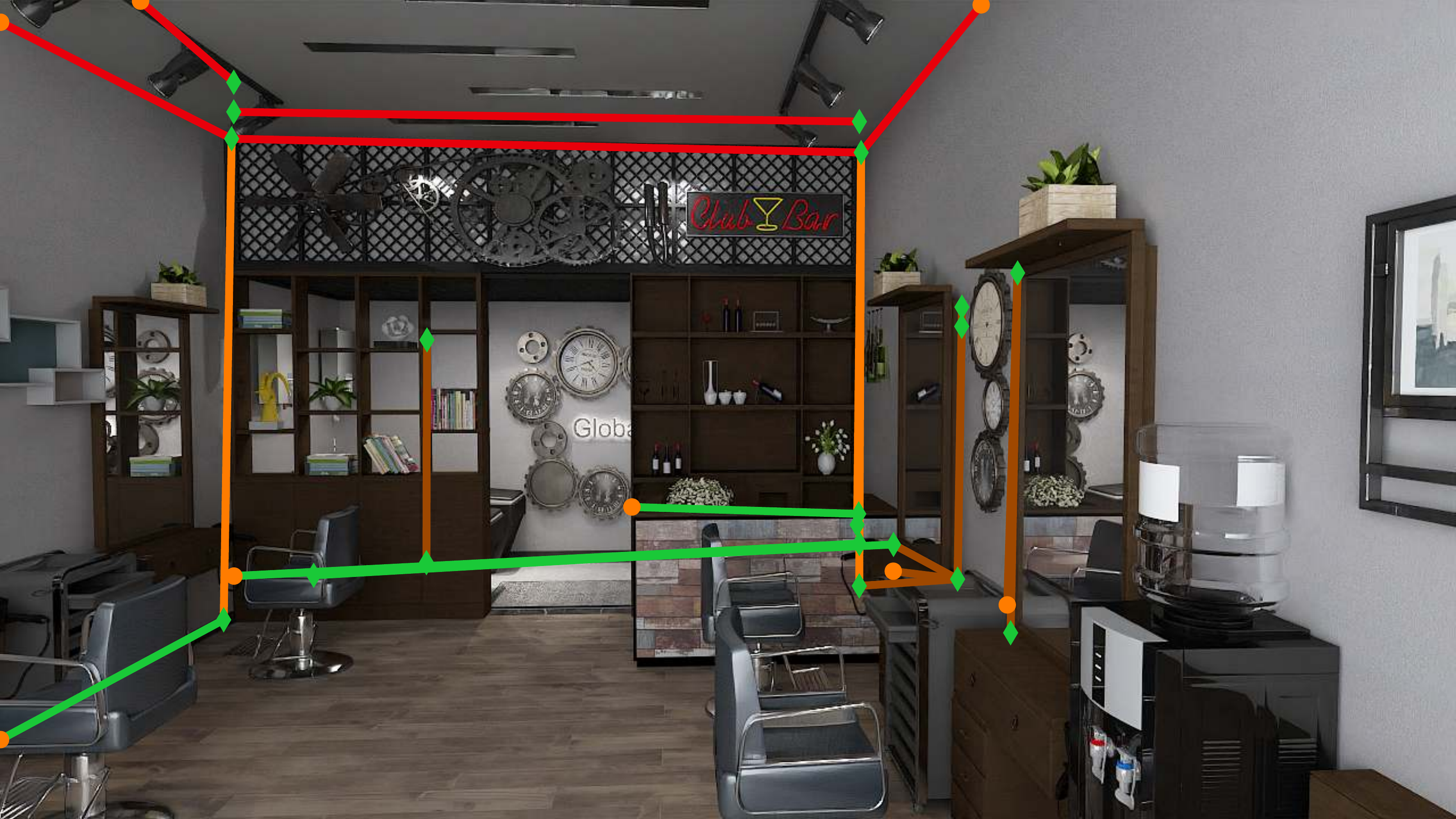}
  \includegraphics[width=\figwidth]{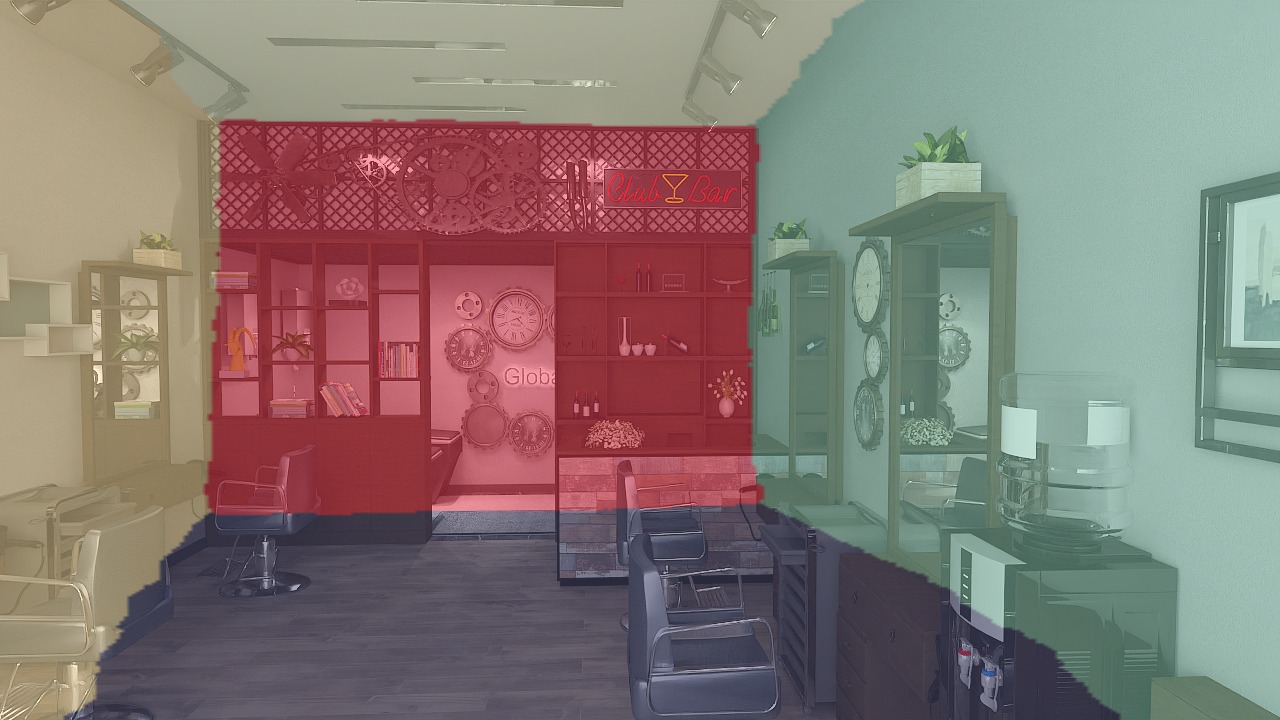}\\

  \includegraphics[width=\figwidth]{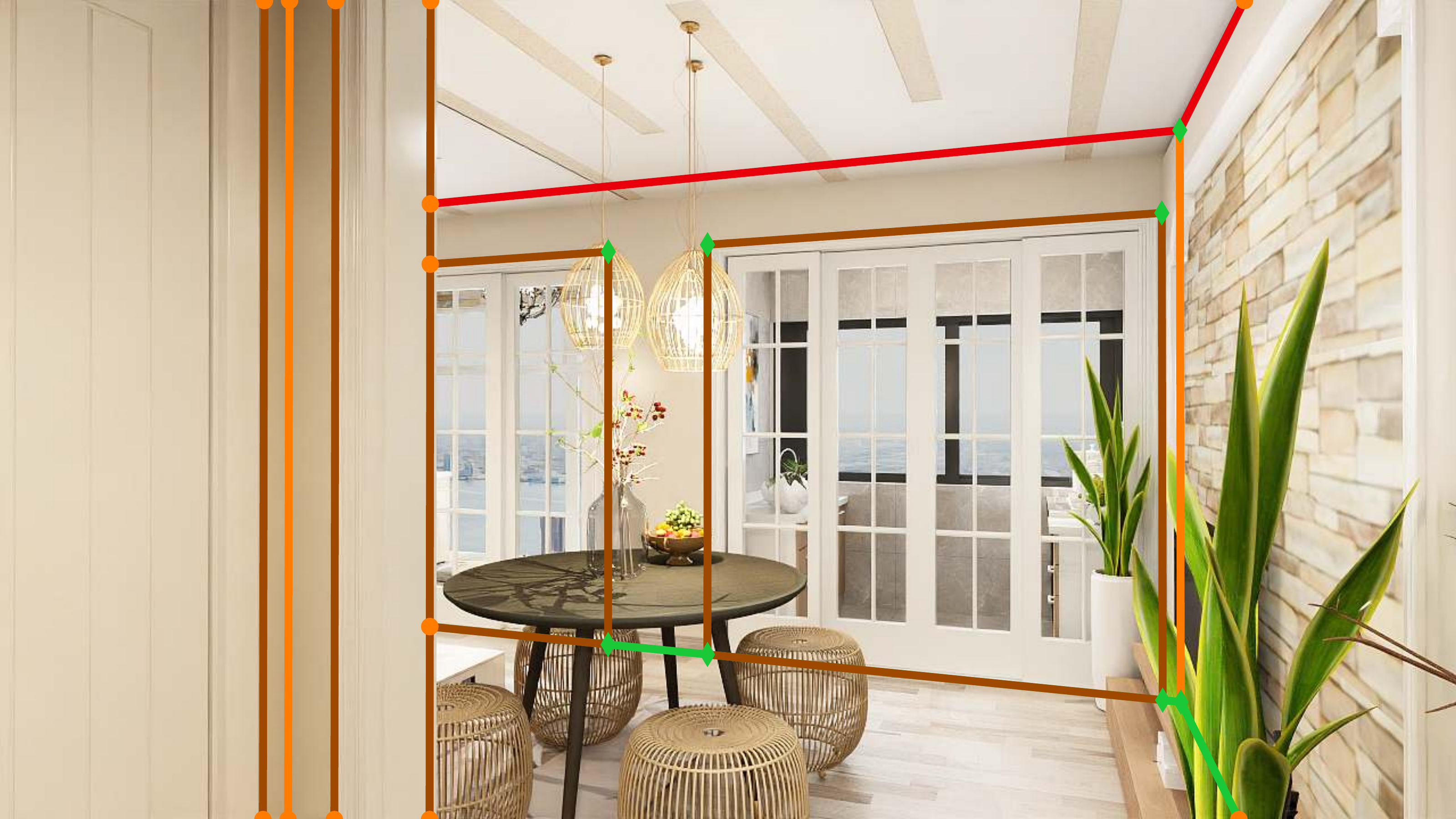}
  \includegraphics[width=\figwidth]{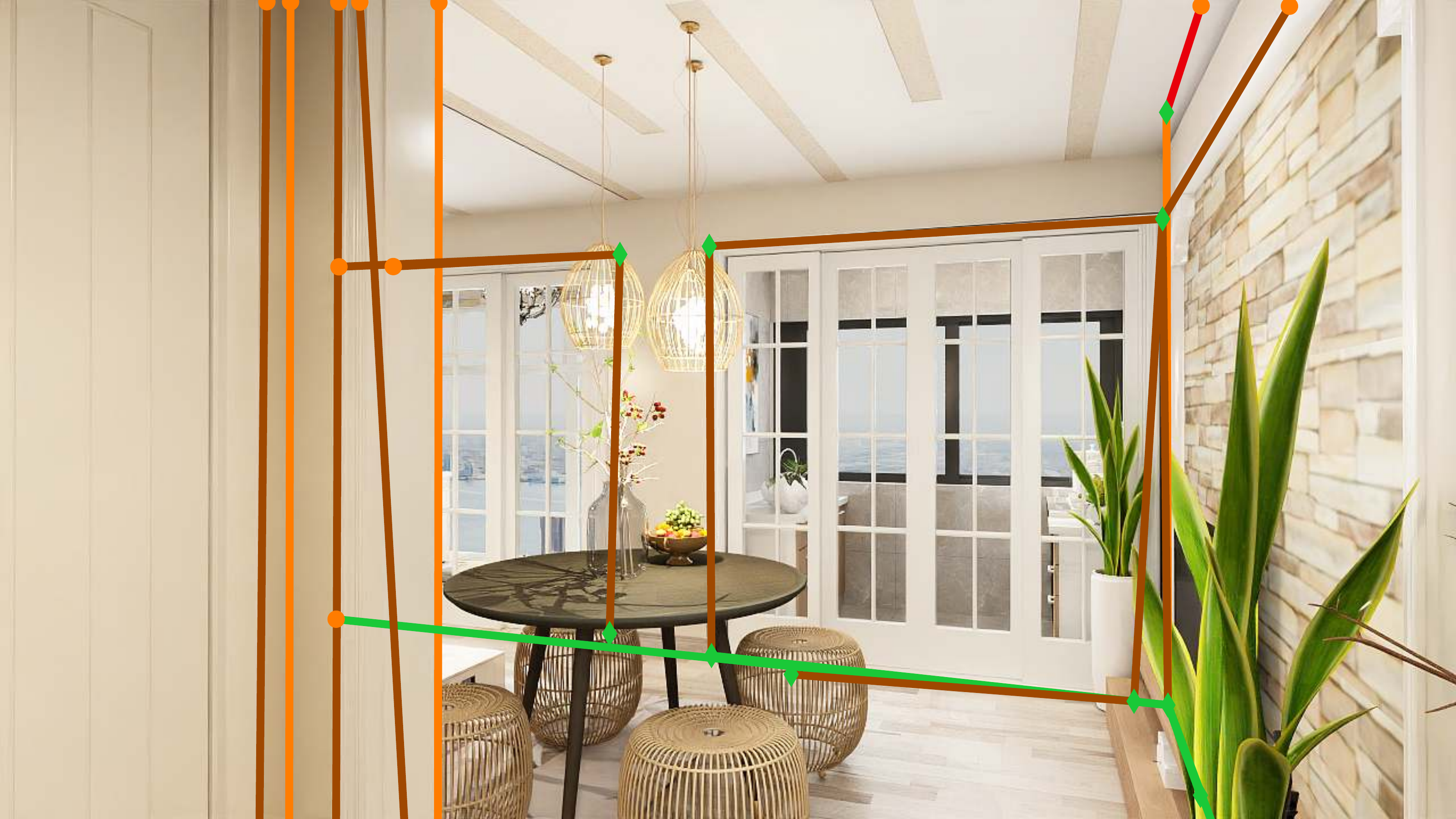}
  \includegraphics[width=\figwidth]{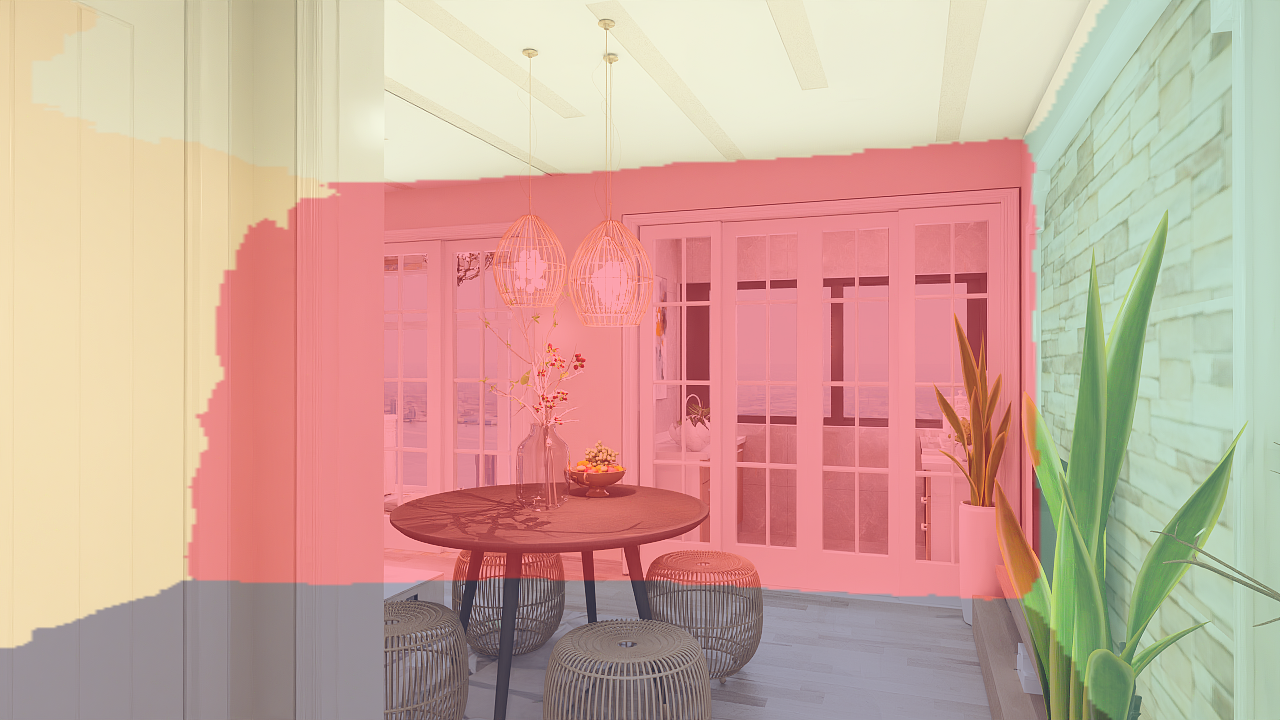}\\

  \includegraphics[width=\figwidth]{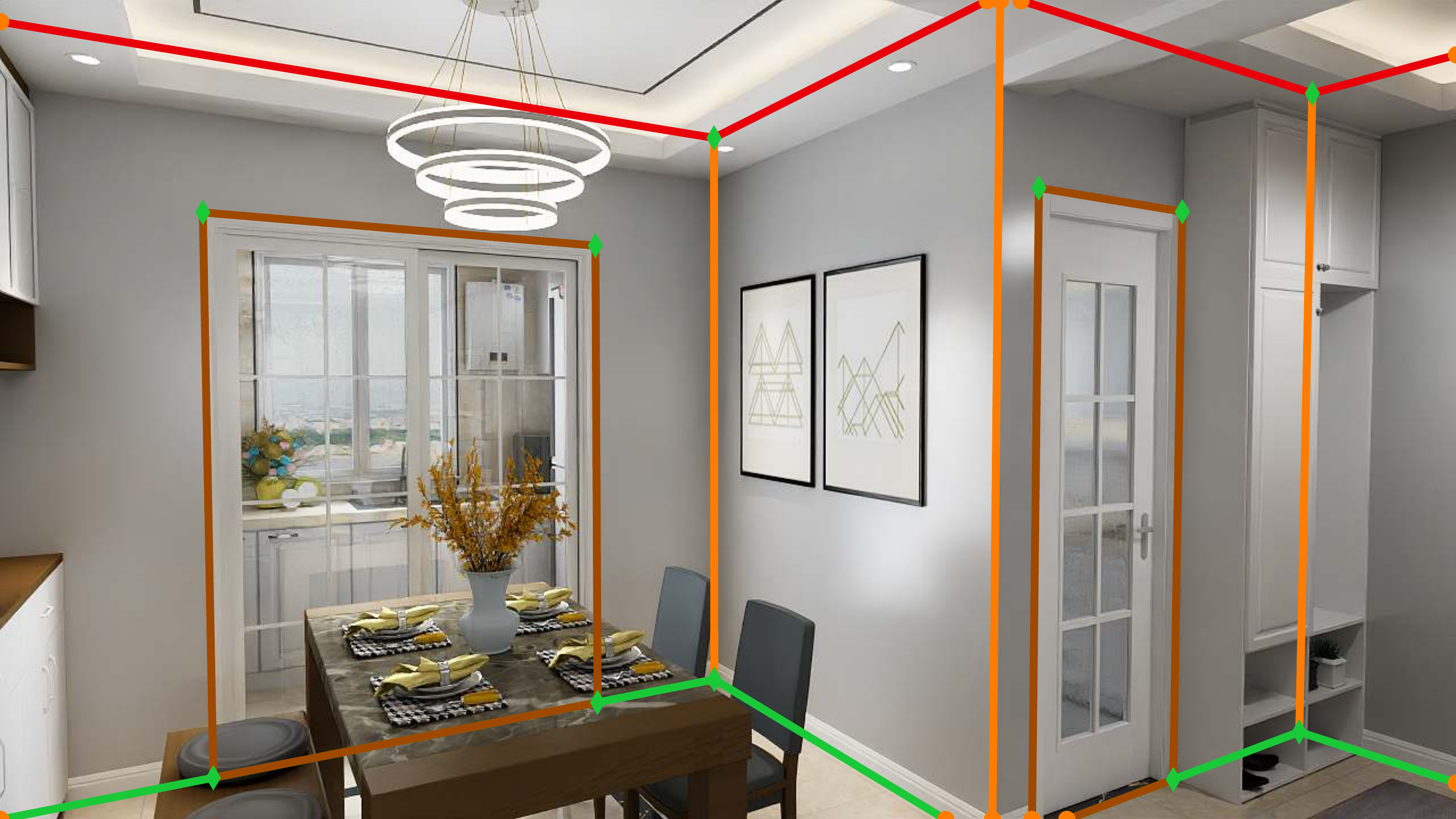}
  \includegraphics[width=\figwidth]{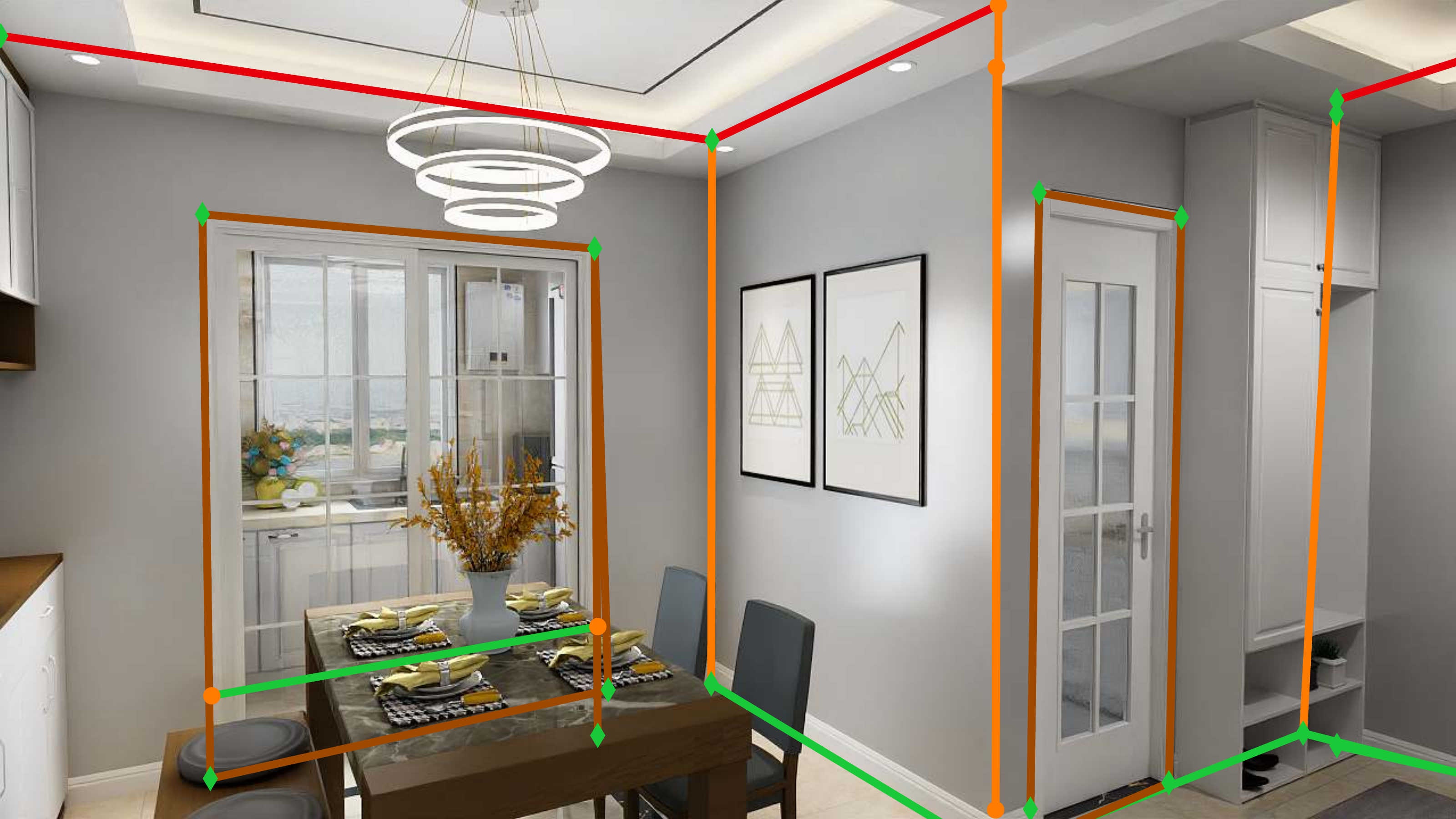}
  \includegraphics[width=\figwidth]{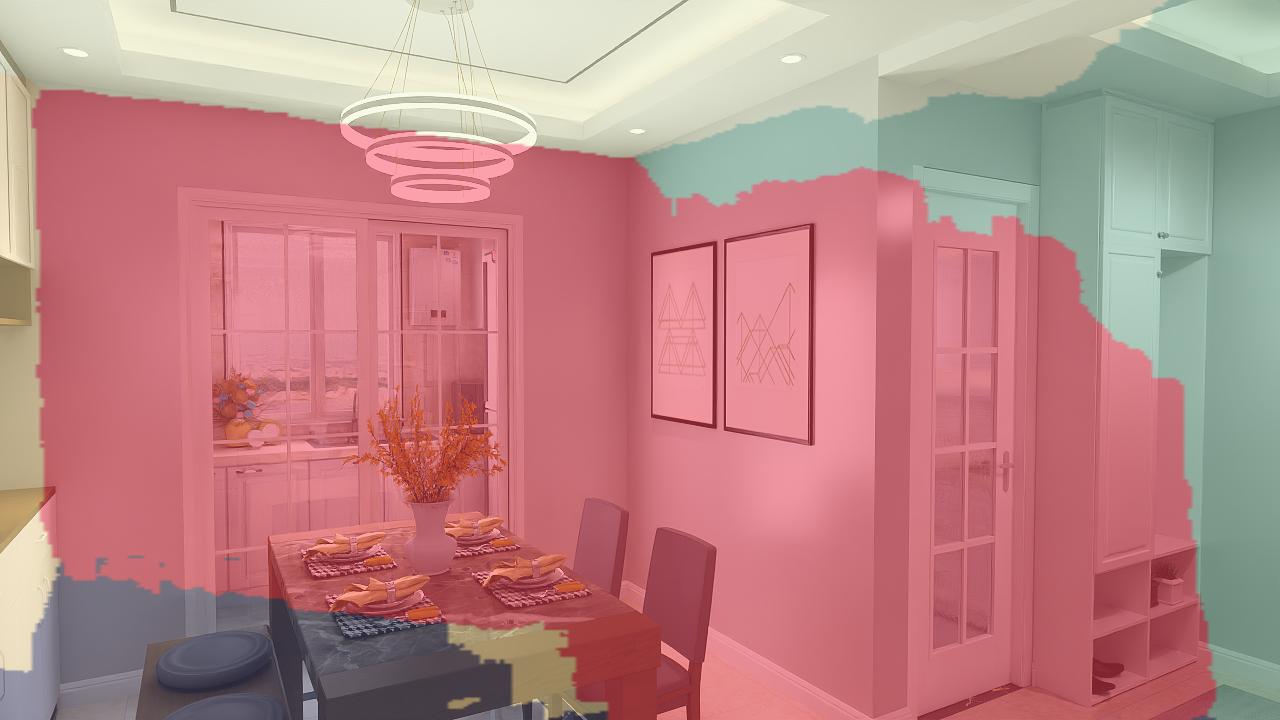}\\

  \includegraphics[width=\figwidth]{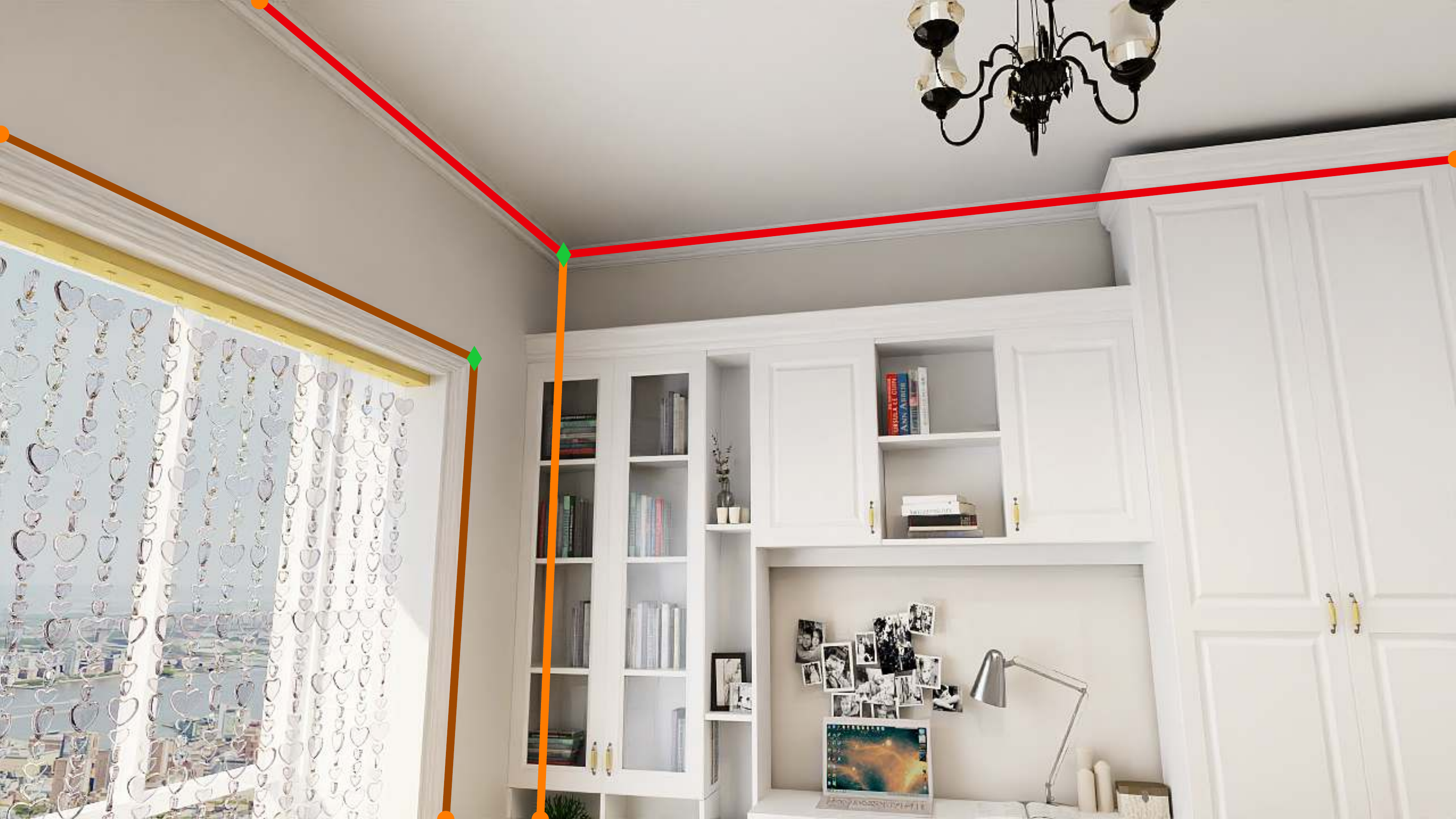}
  \includegraphics[width=\figwidth]{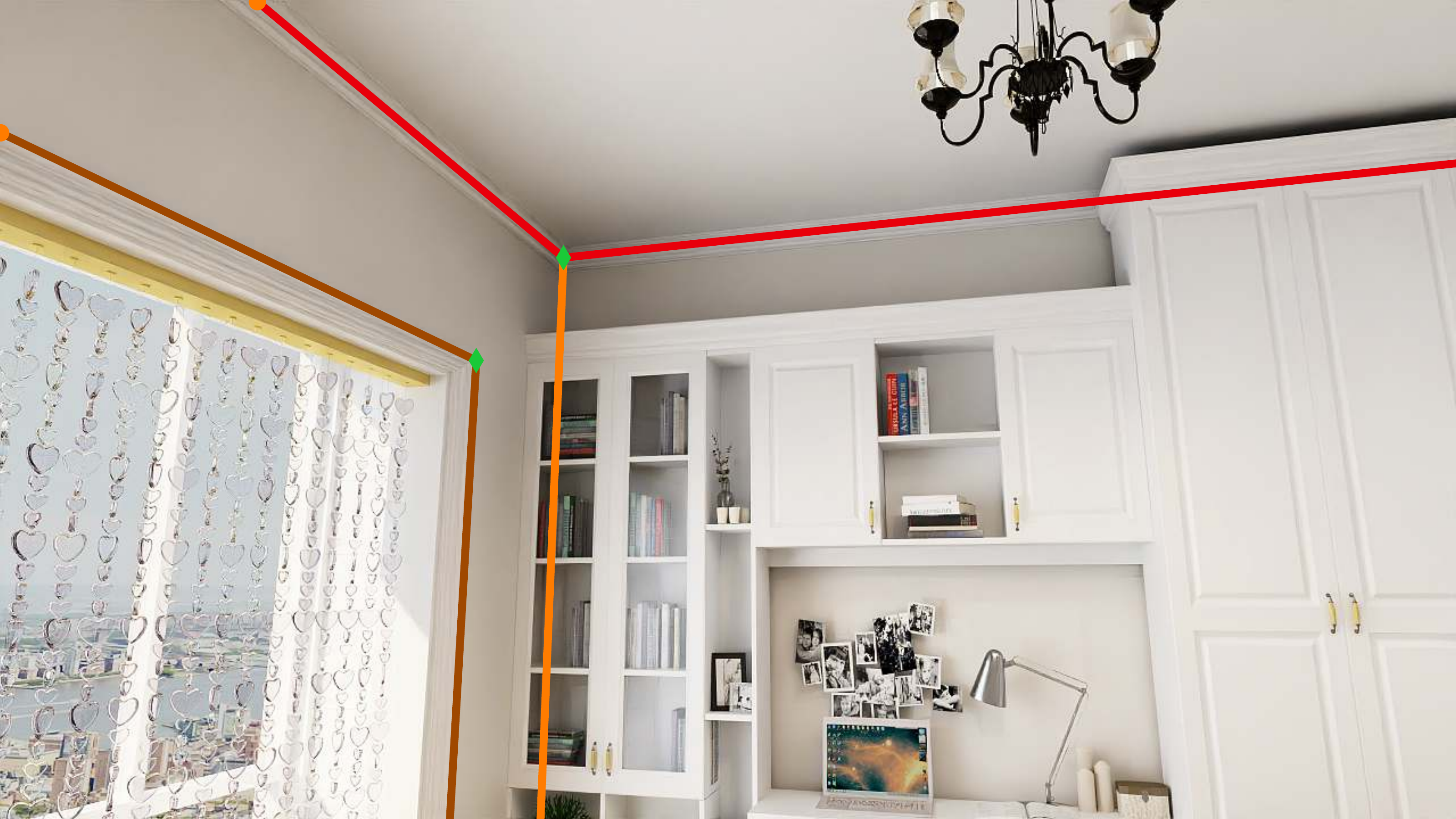}
  \includegraphics[width=\figwidth]{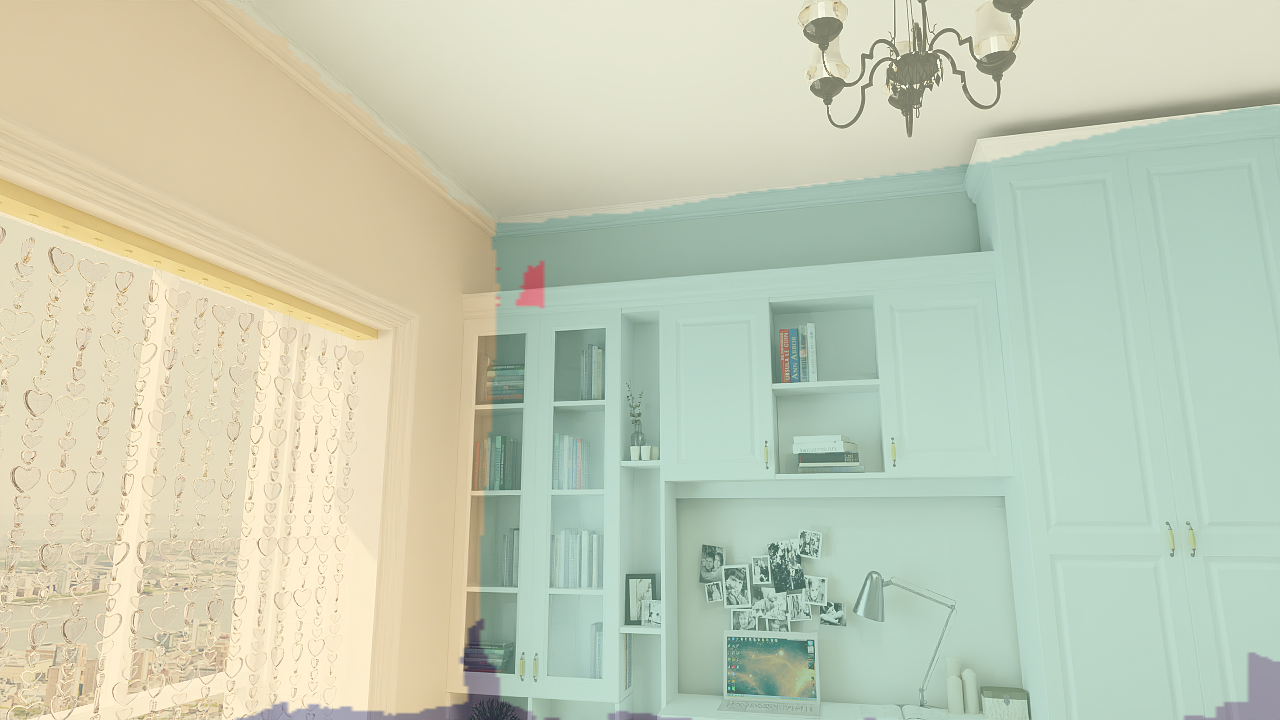}\\

	\caption{Results from the comparison with room layout estimation. On the left: images with ground truth semantic wireframe annotations. In the middle: the results from our method. To the right: room layout results from \cite{lin2018layoutestimation}.}
	\label{fig:roomlayout_vs_semwireframe2}
\end{figure*}

\end{document}